\documentclass[twoside,11pt]{article}

\usepackage{jmlr2e} 
\usepackage[utf8]{inputenc}
\usepackage{amsmath, amssymb}
\usepackage{amsthm}
\usepackage{fancyhdr}
\usepackage{caption}
\usepackage[ruled]{algorithm}
\usepackage{algorithmic}
\usepackage{qtree}
\usepackage{color}
\usepackage{multirow}
\usepackage{colortbl}
\usepackage[table]{xcolor}

\newcommand{\true}{\texttt{true}}
\newcommand{\false}{\texttt{false}}
\newcommand{\ttt}{\texttt{t}}
\newcommand{\fff}{\texttt{f}}
\newcommand{\set}[1]{\ensuremath{\mathbf{#1}}}

\newcommand{\ci}[3]{\ensuremath{I({#1},{#2} \mid #3})}
\newcommand{\cd}[3]{\ensuremath{ \lnot I({#1},{#2} \mid #3})}
\newcommand{\triplet}[3]{\ensuremath{({#1},{#2} \mid #3})}
\newcommand{\blanket}[1]{\ensuremath{\set{B}^{#1}}}

\newcommand{\parameters}{\ensuremath{\mathbf{\Theta}}}
\newcommand{\data}{\ensuremath{\mathcal{D}}}
\newcommand{\closure}{\ensuremath{\mathcal{C}}}

\newcommand{\He}[1]{H_{\textsc{E}}^\star(G_{\tiny{\mbox{#1}}})}
\newcommand{\Hitrue}[1]{{\widehat{H}_{\textsc{I}}}^\star(G_{\tiny{\mbox{#1}}})}
\newcommand{\Hidata}[1]{{\widehat{H}_{\textsc{I}}}^{\mathcal{D}}(G_{\tiny{\mbox{#1}}})}

\newcommand{\rHe}[1]{r^{\star}_{\mathtt{E}}(#1)}
\newcommand{\rHitrue}[1]{r^{\star}_{\mathtt{I}}(#1)}
\newcommand{\rHidata}[1]{r^{\mathcal{D}}_{\mathtt{I}}(#1)}

\newcommand{\gstar}{G^\star}

\newcommand{\ie}{i.e.}
\newcommand{\eg}{e.g.}

\newtheorem{theorem}{Theorem}
\newtheorem*{theorem*}{Theorem}
\newtheorem*{corollary*}{Corollary}
\newtheorem{definition}[theorem]{Definition}

\newtheorem{lemma}[theorem]{Lemma}
\newtheorem{auxiliarylemma}[theorem]{Auxiliary Lemma}

\newtheorem{example}{Example}

\begin{document}

\title{Efficient Independence-Based MAP Approach for Robust Markov Networks Structure Discovery}

\author{\name Facundo Bromberg \email fbromberg@frm.utn.edu.ar\\
\name Federico Schl\"uter \email federico.schluter@frm.utn.edu.ar\\ 
\addr Information Systems Department \\
National Technological University, Facultad Regional Mendoza\\
Rodriguez 273, CP 5500, Mendoza, Argentina}

\editor{}

\maketitle

\begin{abstract}
This work introduces the \textbf{IB-score}, a family of independence-based score functions for robust learning of
Markov networks independence structures. Markov networks are a widely used graphical representation of
probability distributions, with many applications in several fields of science. The main advantage of the
IB-score is the possibility of computing it without the need of estimation of the numerical parameters, an
 NP-hard problem, usually solved through an approximate, data-intensive, iterative optimization. We 
 derive a formal expression for the IB-score from first principles, mainly \emph{maximum a posteriori} and
 conditional independence properties, and exemplify several instantiations of it, resulting in two novel
 algorithms for structure learning: \emph{IBMAP-HC} and \emph{IBMAP-TS}. Experimental results over both
 artificial and real world data show these algorithms achieve important error
 reductions in the learnt structures when compared with the state-of-the-art independence-based structure
 learning algorithm GSMN, achieving increments of more than $50\%$ in the amount of
 independencies they encode correctly, and in some cases, learning correctly over $90\%$ of the edges that GSMN learnt incorrectly.
  Theoretical analysis shows IBMAP-HC proceeds efficiently, achieving these improvements in a time polynomial
  to the number of random variables in the domain.
\end{abstract}


\section{Introduction} 



The present work presents a novel approach for the problem of learning the independence structure of a
Markov network (\textbf{MN}) from data by taking a \emph{maximum-a-posteriori} (\textbf{MAP}) Bayesian perspective of the
independence-based approach \citep{Spirtes00}. Independence-based algorithms learn the structure by performing a
succession of statistical tests of independence among different groups of random variables.
They are appealing for two important reasons: they are amenable to proof of
correctness (under assumptions, see more details on these and other Markov network
concepts in the next section below); and they are efficient, reaching time 
complexities of $O(n^2)$ in the worst case, with $n$ the number of random variables in
the domain. This runtime is an overwhelming improvement over score-based approaches such as maximum-likelihood, that has
super-exponential runtime (more later). 

Unfortunately, the above holds only in theory, as the correctness of the algorithms,
\ie, the guarantee they produce the correct structure, is compromised when the
independence decision of the statistical tests is unreliable, an almost certainty when
these tests are performed on data. Moreover, as we discuss in more detail in the
following sections, for tests to be reliable they require data with size exponential in the number 
of variables $n$.To address these important concerns, the present work focuses on algorithms for improving the quality of independence-based
approaches, while maintaining a manegeable runtime.

Our main contribution is the \emph{IB-score}, the result of a hybrid approach between independence and score based approaches, 
with improvements over both. On one hand, generalizing on
previous works, e.g. \citep{margaritisBromberg09}, it improves the quality of existing independence-based
algorithms by taking a Bayesian, MAP approach, modelling
explicitly the posterior distribution of structures given the data. On the other, it can
be computed  efficiently, contrary to existing scores, as it does not require estimating
the model parameters, an NP-hard computation usually approximated through a
data-intensive iterative algorithm. This is achieved by first  modelling the uncertainty
in the independencies as random variables, and then expanding the posterior over these
independencies, an operation that results in an expression for the posterior of the structures
dependent only on the posterior of independence assertions, that can be computed efficiently (c.f. \S \ref{sec:approach}).

We now proceed in the next Section \ref{sec:graphicalmodels} to discuss in more detail
many of these concepts and algorithms, including a thorough literature review. Following,
Section \ref{sec:approach} formalizes and derives the \textbf{IB-score} from
first-principles. Section \ref{sec:algorithms} introduces two algorithms \textbf{IBMAP-HC}
and \textbf{IBMAP-TS}, each exemplifying two different possible optimization searches for
the MAP, and two different possible instantiations of the IB-score. Then, in Section
\ref{sec:experiments} we present experimental results that confirm the
robustness of IBMAP-HC and IBMAP-TS, and the polynomial runtime of IBMAP-HC. To conclude,
we present a summary and possible directions of future works in Section \ref{sec:conclusions}.

\section{Markov networks}
\label{sec:graphicalmodels}

In this section we describe and motivate MNs in more detail, comparing various existing approaches for learning
an independence structure from data (Section \ref{sec:literature}), expanding on the  deficiencies of the
approaches described above, and how our contribution addresses these deficiencies
successfully.

The innovation rate of computing and digital storage capacity is increasing rapidly as
time passes, producing an important increase in the data available in digital format. As
a response,  the community of data mining and machine learning is constantly producing
novel and improved algorithms for extraction of the knowledge and information implicit in
this data.  Without a doubt, the well-established probabilistic theory is a powerful
modeling tool contributing to  most of these algorithms, specially when the data is
uncertain (\ie, measurements are noisy). \emph{Probabilistic graphical models}
\citep{pearl88, LAURITZEN96, KOLLER&FRIEDMAN09} are a family of multi-variate, efficient
probability distributions that, by codifying implicitly the conditional independences
among the random variables in a domain, produce sometimes exponential reductions in
storage requirements and time-complexity of statistical inference (more details below).
These efficiencies are the reason 
that probabilistic graphical models (or simply graphical models) has had an increasingly
important role in the analysis and design of machine learning algorithms in recent years.

Examples of successful applications include: computer vision \citep{BESAG91,
ANGUELOV&TASKAR05}, restoration of noisy images, texture classification and image
segmentation; genetic research and disease diagnosis \citep{friedman00}; spatial data
mining such as geography, transport, ecology, and many others \citep{shekhar04}.

\begin{figure}[h] \begin{center} \begin{tabular}{c}
\includegraphics[width=\textwidth-1cm]{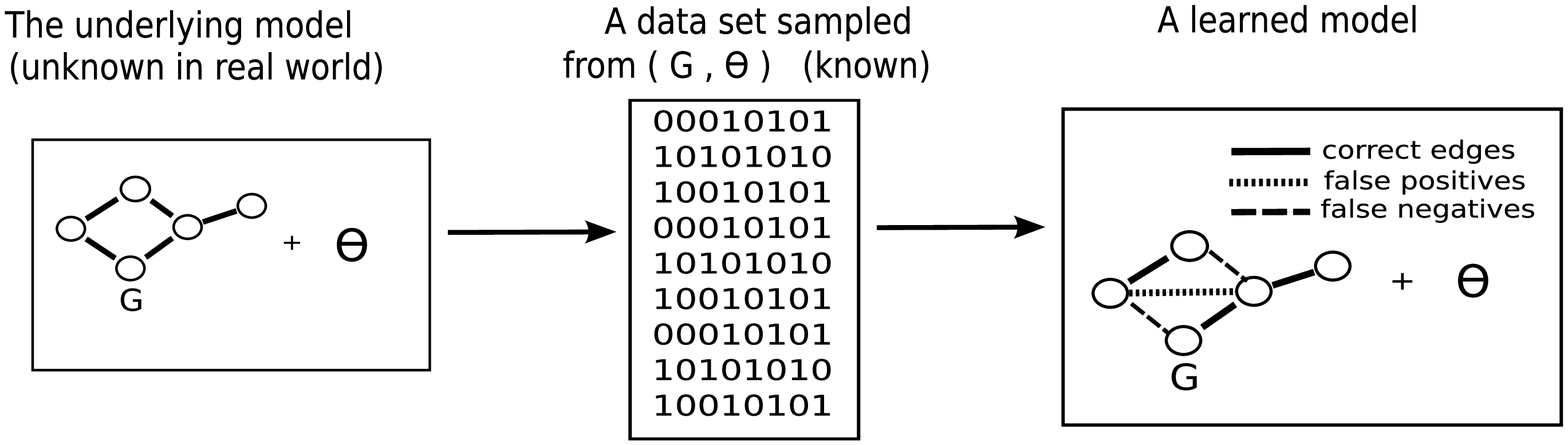}
\end{tabular}
\end{center} 
\caption{  \label{fig:graphicalmodel} Outline of the problem: we have at our disposal a
dataset (center), assumed to be a sampling of some unknown underlying probability
distribution (left). Learning consists in analyzing the input dataset to produce a model
distribution (left). Learning consists in analyzing the input dataset to produce a model
that best fits it  (right), with the hope that it matches the underlying model.} 
\end{figure}

 A graphical model is a probabilistic model over the joint set $\set{V}$ of \emph{random
variables}. It consists in a set of numerical parameters
$\parameters$, and a graph $G$ which encodes ---compactly--- independences among all the
random variables of the domain. The edges of $G$ represent explicitly all the possible
probabilistic conditional independences among the variables, thus called sometimes the
\emph{independence structure} or simply \emph{structure} of the domain. We are interested
in the problem of learning a graphical model from data, that consists in learning both
the independence structure $G$, and the numerical parameters $\parameters$, as
illustrated in Fig.~\ref{fig:graphicalmodel}. As in any statistical modeling process, we
assume the data available (Fig.~\ref{fig:graphicalmodel}, center) is a sampling of some
\emph{unknown} underlying probability 
distribution (Fig.~\ref{fig:graphicalmodel}, left), and learning consists in producing a
structure $G$ and parameters $\parameters$ (Fig.~\ref{fig:graphicalmodel}, right) that
best fit the data, with the hope that this model matches the underlying distribution.

There are two predominant types of graphical models: \textit{Bayesian networks}
(\textbf{BN}) and \textit{Markov networks} (\textbf{MN}). For BNs, $G$ is a directed and
acyclic graph, and for MNs, $G$ is an undirected graph. These two families differ in the
independencies they can encode. Not all sets of independences ---and therefore not all
the probability distributions--- are representable by graphs, and moreover, some sets of
independences ---and therefore some distributions--- are only representable by a BN, and
others only by a MN (and some exists that cannot be represented by neither a BN nor MN).
For the case of MNs (our focus in this work), distributions that can be \emph{faithfully}
representable by undirected graphs are called \textit{graph-isomorph} (see
\citep{pearl88} for more details).   A common practice, and one we follow in this work,
is to assume that the underlying distribution (of which the input data is a sample) is
graph-isomorph.

\begin{figure}[h]
\begin{center}
\begin{tabular}{c}
	\includegraphics[height=3cm]{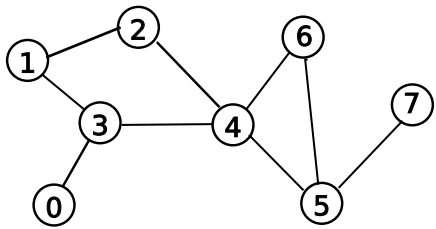} 
\end{tabular}
\caption{ \label{fig:MNexample} An example Markov network with $7$ nodes.}
\end{center} 
\end{figure}

In what follows, we use capitals $X$, $Y$, $Z$, $\ldots$ to denote random variables, and bold
capitals $\set{X}$, $\set{Y}$, $\set{Z}$, $\ldots$ to denote sets of random variables. We
denote by $\set{V}$ the set of all the random variables in the problem domain, and its
cardinality by $n$, \ie, $\mid\set{V}\mid=n$. In this work, we assume that all
variables in $\set{V}$ are discrete. We denote the input dataset by
$\mathcal{D}$, and its cardinality (\ie, the number of data points) by  $N$.   As we
said before, a MN consists in a set of numerical parameters $\parameters$, and a graph
$G$ that encodes conditional independences among the random variables contained in the
set $\set{V}$. We use the notation $\ci{X}{Y}{\set{Z}}$ to denote the
conditional independence predicate $I$ over the triplet $\triplet{X}{Y}{\set{Z}}$ which
is $\true$ ($\false$) whenever $X$ is independent (dependent) of $Y$, given the set of
variables $\set{Z}$.  The graph $G$ consists of $n$ nodes, one for each $X \in \set{V}$,
and a set of edges $E(G)$ between this nodes that encode the set of all probabilistic
conditional independences among the random variables in $\set{V}$.
Fig.~\ref{fig:MNexample} shows an example of a MN. The encoding of independences in a MN is as follows:

\begin{eqnarray}
(X,Y) \in E(G) & \Leftrightarrow & \forall \set{Z} \subseteq \{\set{V} - \{ X , Y \} \},
\cd{X}{Y}{\set{Z}} \label{eqn:encoding}   
\end{eqnarray}

In other words, an edge between two variables $X$ and $Y$ encodes the fact that $X$ 
is dependent of $Y$,  conditioned on any set $\set{Z} \subseteq \{\set{V} - \{ X , Y \} \}$, 
and the lack of an edge between variables $X$ and $Y$ means that 
$\exists \set{Z}   \subseteq \{ \set{V} - \{X,Y\} \}$ that satisfies $\ci{X}{Y}{\set{Z}}$.

As shown in \cite{pearl88}, when the underlying model is graph-isomorph, the encoding
is equivalent to reading independences through \emph{vertex-separation}, \ie, two
variables $X$ and $Y$ are independent (dependent) according to the graph given the set
$\set{Z}$, if and only if variables $X$ and $Y$ are disconnected (connected) in the
sub-graph of $G$ resulting from removing all edges incident in a variable in $\set{Z}$. 
In other words, the set $\set{Z}$ intercepts (does not intercepts) every path from $X$
to $Y$. To illustrate, consider variables $0$ and $5$. These variables are dependent
given $\{2\}$ because there is still a path through variable $3$, and thus independent
given $\{2, 3\}$.

As mentioned, the strength of graphical models, both MNs and BNs, lies in the sometimes
exponential reduction in storage capacity and time complexity of statistical inference.
Storage and statistical inference (mostly marginalization) is exponential because the
representation of a multi-variate joint probability distribution over $\set{V}$ consists
in a multi-dimensional table with $n+1$ columns, and exponentially
many tuples, each consisting in a complete assignments of the $n$ variables, and one
column to store the probability of that assignment.  Fig.~\ref{fig:mutually} (left)
illustrates the table for the joint probability of a system of $n$ binary variables
(\ie, that can take any of two values), consisting on $2^n$ tuples, one per
configuration. The exponential
reduction in storage and inference occurs because certain conditional independences
allow a decomposition of the  joint probability into a product of \emph{factor} or
\emph{potential} functions, each quantified by the parameters $\parameters$, and
dependent on only a subset of $\set{V}$. For instance, it is a well-known fact that when
all $n$ variables in $\set{V} = \{V_1,...,V_n\}$ are mutually independent,
$\Pr(V_1,...,V_n) = \prod\limits^n_{i=1} {\Pr(V_i)}$.  Since the tables associated with a
function over some set of  variables is exponential in the number of variables in that
set (as exemplified earlier in this paragraph for the case of the joint), the
decomposition requires then a polynomial number of (exponentially) smaller tables.
Moreover, it can be shown that in many cases the number of variables in each factors is
bounded by a small number, resulting in the exponential reduction in both storage and
marginalization. To continue with our example, the decomposition of the $n$ binary
variables results in $n$ tables with only $2$ tuples each, as illustrated in
Fig.~\ref{fig:mutually} (right).    An important difference between MNs and BNs lies in
the properties of their factor functions. The factor functions of MNs are not normalized,
thus, in order to obtain a fully quantified probabilistic model, an exponential
computation of a normalization constant (known as the \emph{partition function}) over all
possible assignments of the $n$ variables is required. BNs, instead, allow a
factorization of the joint distribution into conditional probability distributions, i.e.,
normalized factors, that can thus be learned efficiently from data (does not require
normalization). 

\begin{figure}[h]
\begin{center}
\begin{tabular}{c}
	\begin{tabular}{|cccc|c|}
		\hline
	$V_1$ &  $V_2$  & ... & $V_n$  & $ \Pr(V_1,..,V_n)$  \\
		\hline
	0 & 0 & ... & 0 & 0.121 \\
	0 & 0 & ... & 1 & 0.076 \\
	. & . & ... & . & . \\
	. & . & ... & . & . \\
	. & . & ... & . & . \\
	1 & 1 & ... & 0 & 0.21 \\
	1 & 1 & ... & 1 & 0.12 \\
	
		\hline
	\end{tabular}

	$\mathbf{\Rightarrow}$

	\begin{tabular}{|c|c|}
		\hline
	$V_1$ &  $ \Pr(V_1)$  \\
		\hline
	0 & 0.21 \\
	1 & 0.79 \\
		\hline
	\end{tabular} , 
	\begin{tabular}{|c|c|}
		\hline
	$V_2$ &  $ \Pr(V_2)$  \\
		\hline
	0 & 0.45 \\
	1 & 0.55 \\
		\hline
	\end{tabular}
	...  
	\begin{tabular}{|c|c|}
		\hline
	$V_n$ &  $ \Pr(V_n)$  \\
		\hline
	0 & 0.42 \\
	1 & 0.58 \\
		\hline
	\end{tabular}

\end{tabular}
\end{center} 

\caption{\label{fig:mutually}Storage reduction over $n$ binary variables from $2^n$ to $2n$.}

\end{figure}
\subsection{Structure learning algorithms: background and related work} 
\label{sec:literature}

In this section we discuss the problem, and motivate our approach for solving it, in the 
context of existing algorithms for learning structure from data.

Historically, the literature has presented two broad approaches for learning MNs: \emph{score-based}
algorithms, and \emph{independence-based} (also called \emph{constrained-based})
algorithms.   Score-based algorithms, exemplified by
\cite{LamBacchus94,MCCALLUM03,ACID03}, perform a search on the space of all possible
graphs to find the structure with maximum score. These algorithms differ in the approach
taken to explore the space of all graphs (super-exponential in size, \eg, there  are $2^{(n~choose~2)}$
undirected graphs with $n$ nodes), and in the functional form, and theoretical
justification, for the score. Examples of scores are \emph{maximum likelihood} $Pr(\data
\mid G, \parameters)$ of the data $\data$ given the model $(\parameters, G)$
\citep{BESAG74}, \emph{minimum description length} of the model $(\parameters, G)$
\citep{LamBacchus94}, or pseudo-likelihood \citep{BESAG75,YU-CHENG2003,
KOLLER&FRIEDMAN09}), again of the 
$\data$ given the model $(\parameters, G)$. The last approach, pseudo-likelihood,
introduces an approximate expression for computing the likelihood that does not require
the normalization of the model (\ie, the computation of its normalization constant).  To
compute the score of each structure $G$ ``visited'' during the search, it is
necessary to estimate the parameters $\parameters$. For BNs this operation can be
performed efficiently \citep{Heckerman95}, but for MNs this is a costly operation that
consists in an optimization in itself \citep{BESAG74, YU-CHENG2003, KOLLER&FRIEDMAN09}.
For instance, in the 
case of the likelihood, the  optimization would be over the  space of parameters, for a
fixed structure $G$. We will later show that our approach, although it  requires
a search over the space of structures, can compute efficiently the (independence-based) score at each step
as it does not require the estimation of the parameters.
 
 Independence-based algorithms \citep{Spirtes00, HITON03, IAMB03, bromberg&margaritis09b,
 margaritisBromberg09} have the ability to learn the independence structure efficiently,
 as they do not require neither a search over structures, nor the estimation of
 parameters at each step of their execution (of course, in order to obtain a complete
 model $(G, \parameters)$ they do need to estimate the parameters once the structure has
 been learned).  Instead, these algorithms take a rather direct approach for
 constructing the independence structure by inquiring about the conditional independences
 that hold in the input data. This inquiry is performed in practice through statistical
 tests such as Pearson's $\chi^2$ test \citep{AGRESTI02}, more recently the \emph{Bayesian
 test} \citep{MARGARITIS05,margaritisBromberg09}, and for continuous Gaussian data the
 \emph{partial correlation} test \citep{Spirtes00}, among others. At each step, these algorithms
 propose a triplet $\triplet{X}{Y}{\set{Z}}$ and inquire the data whether independence
 or dependence holds for that triplet. Which triplet is proposed at some step depends on
 the independence information the algorithm has up to that point, that is, which triplets
 has been proposed so far, and their corresponding independence value. To find the
 structure consistent with this independence information, these algorithms proceed by
 discarting, at each step, all those structures that are inconsistent (\ie, do not
 encode) the independence just learned, until all but one structure has been discarded.
 Algorithms exist that require a number of tests that is polynomial in the number $n$ of
 variables in the domain, which, together with the fact that statistical tests can be
 executed in a running time proportional to the number of data points $N$ in the dataset,
 result in a total running time that is polynomial in both $n$ and $N$. On top of their
 efficiency, another important advantage of these algorithms, is that, under assumptions
 (statistical tests are correct, and the underlying model is graph-isomorph), it is
 possible to prove the correctness of these algorithms, \ie, that they return the
 structure of the underlying model. For a thorough example of such proof we refer the
 reader to \cite{bromberg&margaritis09b}. Unfortunately, statistical independence tests
 are not always reliable. Most independence-based algorithms are oblivious to this fact,
 evident from their design, which discards structures based on a single test. If the test
 is wrong (\ie, it asserts independence when in fact the  variables in the triplet are
 dependent, or vice versa), the underlying (true) structure may be discarded. This
 problem is not to be underestimated because the quality of statistical independence
 tests degrades exponentially with the number of variables involved in the test. For
 example, \citep{COCHRAN54} recommends that Pearson's $\chi^2$ independence test be
 deemed unreliable if more than 20\% of the cells of the tests contingency table have an
 expected count of less than 5 data points. Since the contingency table of a conditional
 independence test over $\triplet{X}{Y}{\set{Z}}$  is $d$-dimensional, with $d = 2 + \mid
 \set{Z}\mid$ \citep{AGRESTI02}, the number of cells, and thus the number of data points
 required, grows exponentially with the size of the test. In other words, for a fixed
 size dataset, the quality of the test degrades exponentially with the size of its
 conditioning set.

 This problem has been addressed by \cite{BrombMarg09} using Argumentation. Their
 approach modeled the problem as an inconsistent knowledge base consisting on Pearl's
 axioms of conditional independence as rules, and the triplets and their corresponding
 assignments of independence values as fact predicates. If the underlying model is
 graph-isomorph, its set of independencies must satisfy the axioms (see \cite{pearl88,
 KOLLER&FRIEDMAN09} for details). In practice, however, independencies are not queried
 directly to the underlying model, but to its sampled dataset. These independencies may
 be unreliable, and thus may violate the rules. Argumentation is an inference procedure
 designed to work in inconsistent knowledge bases. \cite{BrombMarg09} used this
 framework to infer independencies, sometimes resulting in inferred values different
 than measured values. Taking the inferred value as the correct one, they obtained
 important improvements in the quality of the reliability of statistical independence
 tests, and thus of the independence structures discovered.
 
 We present here an alternative approach to the unreliability problem. Our approach is
 inspired by the work of \cite{margaritisBromberg09},  
 that designed an algorithm that instead of discarding structures based on outcomes of
 statistical tests, maintains a distribution over structures conditioned on the data,
 \ie, the posterior distribution of the structures. To learn the structure, the
 algorithm takes the maximum-a-posteriori approach. In the next section we present the
 \emph{IB-score} (independence-based score), an expression for computing efficiently the
 posterior of some structure based on outcomes of statistical independence tests, and in
 the following sections we use this score in two different algorithms that conduct the
 search for its maximum. The resulting algorithms are hybrids of score and independence
 based algorithms. Score-based because they proceed by maximizing a score, \ie, the
 posterior of the structures given the data or as we call it, the IB-score; and
 independence-based, because the IB-score  is computed through statistical tests of
 independence (thus its name). However, contrary to previous score-based approaches, the
 computation of the score (the IB-score) does not require the estimation of the numerical
 parameters $\parameters$ and is therefore efficient.

\section{Our approach: independence-based MAP structure learning} 
\label{sec:approach}

As discussed in the previous section, independence-based algorithms has the advantage of
not requiring neither a search over structures, nor an interleaved estimation of the
parameters estimation at each iteration of the search (a search in itself,
\citep{YU-CHENG2003}). Unfortunately, these advantages are compensated by a major
robustness problem as the vast majority of the independence-based algorithms rely blindly
on the correctness of the tests they perform, with the risk of discarting the true,
underlying structure when the test is incorrect.  This problem is exacerbated
by the fact that the reliability of statistical independence tests degrades exponentially
with the size of the conditioning set  of the test (for some fixed size dataset). To
overcome this robustness problem we take a Bayesian approach, inspired and extended from
\citep{MARGARITIS05,margaritisBromberg09}. This approach models the problem of MN
structure learning as a distribution over structure given the data, \ie, the posterior
distribution of structures. Under this model, structures that are inconsistent with the
outcome of a test are not discarded, but their probability is reduced. The approach taken
is a combination of score and independence based algorithms for structure learning.
Score-based because we search for the structure $G^{\star}$ whose posterior probability
$\Pr(G \mid \data)$ is maximal, that is, we take the maximum-a-posteriori (\textbf {MAP})
approach:              

\begin{equation} \label{eq:maxIB-score}
	G^{\star} = \arg\max_{G}{\Pr(G \mid \data)},  
\end{equation}

and independence-based because the expression obtained for computing the posterior
probability is based on the outcome of statistical independence tests, as explained in
detail in the following section. Later, in section \ref{sec:algorithms} we present two
practical algorithms for conducting the maximization: \textbf{IBMAP-HC}
(Independence-based maximum a posteriori hill climbing) and \textbf{IBMAP-TS}
(Independence-based maximum a posteriori tree search).     

\subsection{Independence-based posterior for MN structures: the IB-score} 
\label{sec:IB-score}

In this section we present the \emph{IB-score}, a computationally feasible expression for
the posterior $\Pr(G \mid \data)$. We proceed by re-expressing the posterior of 
structure $G$ in terms of the posterior of a particular set of  independence assertions
$\closure(G)$ we call the \emph{closure}, and then, through
approximating assumptions, obtain an efficiently computable expression for the posterior
we call the IB-score.

Let us first define formally the \emph{closure} $\closure(G)$ of a structure $G$:

\begin{definition}[Closure] \label{def:closure}
A \emph{closure} of an undirected independence structure $G$ is a set $\closure(G)$ of 
conditional independence assertions that are sufficient for determining $G$.     
\end{definition}

We can illustrate this definition by a couple of examples:
\begin{example}
According to Eq.~(\ref{eqn:encoding}), we can construct a closure for $G$ by adding the
assertion $\ci{X}{Y}{\set{V}-\{X,Y\}}$ for 
every pair of variables $X, Y \in \set{V}$, $X \neq Y$,  for which there is no edge in
$G$, \ie, $(X,Y) \notin E(G)$. Otherwise, if there is an edge between them, \ie, $(X,Y) \in E(G)$, add the
assertions $\cd{X}{Y}{\set{Z}}$ for every $\set{Z} \subseteq \set{V}-\{X,Y\}$.
\end{example}

\begin{example} \label{ex:algorithm-based-closure}
Our second example considers any independence-based algorithm
for which exists a proof of correctness; under the usual assumptions of
faithfulness and correctness of the independence assertions.  That is, proof that when
provided with correct independence values for each independence inquiry, the structure
$G^{\star}$ output by the algorithm is the only one consistent with those values. 
Being $G^{\star}$ the only possible structure for those independencies, the independencies determine
$G^{\star}$, and they conform a closure. Examples of correct algorithms for MNs are the
GSMN and GSIMN algorithms \citep{bromberg&margaritis09b}.      
\end{example}

We now prove an important Lemma that asserts that the posterior $\Pr(G \mid \data)$ of a
structure exactly matches the posterior $\Pr(\closure(G) \mid  \data)$, for any
closure of $G$ $\closure(G)$.   

\begin{lemma}[Probabilistic equivalence of structures and closures]
For any undirected independence structure $G$ and any closure $\closure(G)$ of $G$, it
holds that 
\[
	\Pr(G \mid \data) = \Pr(\closure(G) \mid  \data).
\]
\end{lemma}
\begin{proofJMLR}
The proof proceeds by incorporating, in two steps, the information of independence
assertions contained in the closure $\closure(G)$ of $G$ into the posterior $\Pr(G \mid \data)$.
First, we model as random variables the uncertainty of independence assertions obtained through
unreliable statistical independence tests. Formally, the
uncertainty in the independence assertion $I(T)=t$, denoting that the triplet $T$ is
independent (dependent) when $t=\true$ ($t=\false$), is formalized by a random variable
$T$ taking the values $t \in \{\true, \false\}$ (note the notation overload). Second, for
each independence assertion $\left(I(T_i)=t_i\right) \in \closure(G)$, $i=1,\ldots,c$,
and $c=|\closure(G)|$, we incorporate $T_i$ (its corresponding random variable) into the
posterior using the \emph{law of total probability}, namely: 

\begin{equation*} 
 	\Pr(G \mid \data) = \sum\limits_{t_1 \in \{\ttt,\fff\}} \sum\limits_{t_2 \in
	\{\ttt,\fff\}} \ldots \sum\limits_{t_c \in \{\ttt,\fff\}} {\Pr(G, T_1=t_1, T_2=t_2,
	\ldots,  T_c = t_c \mid \data )}  
\end{equation*}
where $\ttt$ and $\fff$ are abbreviations of $\true$ and $\false$, respectively. We
simplify this expression by first collapsing the $c$ uni-dimensional summations over
$\{t,f\}$ into a single, multi-dimensional summation over $\{t,f\}^c$, then abbreviating
$T_i=t_i$ by $t_i$, and then $\{t_1, t_2, \ldots, t_c\}$ by $t_{1:c}$, obtaining    

\begin{equation*} 
 	\Pr(G \mid \data) = \sum\limits_{t_{1:c} \in \{\ttt,\fff\}^c }  {\Pr(G, t_ {1:c}  \mid \data )}.
\end{equation*}

Applying the chain rule to the terms in the summation we obtain
\begin{equation*} 
 	\Pr(G \mid \data) = \sum\limits_{t_{1:c} \in \{\ttt,\fff\}^c }  {\Pr(G \mid t_{1:c},
	\data )} \times {\Pr(t_{1:c}  \mid \data )} 
\end{equation*}

Now, by definition of a closure, a structure $G$ is determined by its independence
assertions, \ie, the probability of $G$ given these assertions (and any other variable
such as $\data$), must equal $1$. Moreover, if we flip the independence value of any of
these assertions, then it is clear that $G$ cannot be the structure, in other words, its
probability given the flipped assertions (and any other variable such as $\data$) must be
$0$. This results in the left factor $\Pr(G \mid t_{1:c}, \data )$ in all terms of the
summation being $0$, except for the term containing the assignments consistent with the
closure (where it is $1$). If we denote by $\{t_1^G,t_2^G,\ldots,t_c^G\} = t_{1:c}^G$
these assignments, we get        
\begin{equation*} 
 	\Pr(G \mid \data) =  \Pr(t^G_{1:c}  \mid \data ),
\end{equation*}
and the lemma is proved by noticing that  $t^G_{1:c}$ is not more than the closure $\closure(G)$.
\end{proofJMLR}

Unfortunately, to the best of the authors knowledge, there is no method for computing the
joint $\Pr(\closure(G) \mid \data)$ of many independence assertions. We are forced then to
make the approximation that the independence assertions in the closure are all mutually
independent. This gives us the IB-score($G$) of $G$:
\begin{equation}  \label{eq:independenceApproximation}
 	\widehat\Pr( \closure(G) \mid \data) =  \mbox{IB-Score($G$)} = \prod\limits_{(I(T)=t) \in \closure(G)}  \Pr( T=t \mid \data )
\end{equation}
which can now be computed using the Bayesian test of
independence of \cite{MARGARITIS05, margaritisBromberg09} to  
compute the factors $\Pr( T=t \mid \data )$.

The approximation implies that our belief in the independence of any triplet $T$ in the
closure is not affected by our knowledge that some other triplet $T'$ in the closure is
independent (or dependent). This my not be true in practice. It is certainly false when
the two triplets are related through Pearl axioms of independence \citep{pearl88}.
In that case they determine each other, \ie, $\Pr(T \mid T', \data)=1$, and
thus their joint probability is easily computable: $\Pr(T, T'\mid \data) = \Pr(T\mid
T'\data) \times \Pr(T'\mid \data) = \Pr(T'\mid \data)$. For triplets that are not related
through Pearl axioms we are forced into this approximation until a method for computing
the joint over several independence assertions is developed (if this is at all possible).


To conclude this section we note that we had made a single approximation, namely, that
the random variables of independence assertions are mutually independent. The expression
found can be computed efficiently, with a runtime complexity proportional to the sum of
the complexities of the statistical tests performed for computing the probability $\Pr(T=t
\mid \mathcal{D})$ of each factor. In \cite{BrombMarg09} it is argued that the
computational cost of performing a statistical test on data  
for the triplet $\triplet{X}{Y}{\set{Z}}$
is proportional to the number of data points in the dataset, \ie, $N$, times the total number
of variables involved in the test, \ie, $2 + |\set{Z}|$. So in the worst case,
the complexity of the IB-score would be
\begin{equation} \label{eq:complexityIBScore}
	O(|\closure(G)|.N.\tau^*), 
\end{equation}
where $\tau^*=\max_{T \in \closure(G)} |T|$, \ie, the number of variables in the triplet,
among all triplets in the closure, with the maximum number of variables.

\section{Practical algorithms for MAP optimization}
\label{sec:algorithms}

A full specification of the MAP search of Eq.~(\ref{eq:maxIB-score}) requires the
specification of the search mechanism and the closure chosen for the structures. We
present in what follows two approaches: \textbf{IBMAP-HC} (Independence-based maximum a
posteriori hill climbing) and \textbf{IBMAP-TS} (Independence-based maximum a posteriori
tree search) that differ in both the search mechanism (hill climbing and tree search,
respectively), and in the choice of closure (Markov blanket based, and algorithm based,
respectively). The motivation for these choices is two-fold. First, they serve as more
realistic examples of closures than those given in Section \ref{sec:IB-score}, and
second, the more implementations of the IB-score, more robust the experimental
conclusions. Let us discuss in detail each of these approaches.

\subsection{IBMAP-HC: Independence-based MAP structure learning using hill climbing and
Markov based closure.} 
\label{sec:IBMAP-HC}

We proceed now with the specification of the closure and search procedure chosen for
IBMAP-HC.  

Given a domain of variables $\set{V}$, $n=|\set{V}|$, we propose the hill climbing local
search mechanism for finding the best structure in the space of all undirected structures
of size $n$. Starting from some structure $G$, the search proceeds finding the structure
$G'$ with maximum IB-score among all structure one \emph{edge-flip} away from $G$. An
edge-flip of some pair of variables $(X,Y)$, consists on removing the edge $(X,Y)$ from
$G$, if such edge exists, or adding it otherwise. The amount of neighbors thus equals the
number of pairs of variables, which is $n(n-1)/2$. The algorithm continues recursively
from $G'$, until all structures one flip-away has smaller IB-score, \ie, until the
algorithm reaches a local maxima. As a starting structure we chose the structure output
by the GSMN algorithm \citep{bromberg&margaritis09b}. This way, the hill climbing search
can be seen as a perturbation 
of the output of GSMN, with the hope that the local maxima
in the proximity of GSMN has better quality. Experimental results confirm this is in fact
the case. This could be done of course with structures output by any other structure
learning algorithm, although we expect improvements only on independence-based
algorithms. Our experiments show results for GSMN only.           

For the closure, we chose one based on \emph{Markov blankets}. Let us
first define the concept of Markov blanket and then explain how it can be used to specify
a closure.  

\newtheorem*{definitionStar}{Definition}
\begin{definitionStar}[\citet{pearl88}, p.97]
	The \emph{Markov blanket} of a variable $X \in \set{V}$ is a set $\blanket{X} \subseteq
	\set{V} - \{X\}$ of variables that ``shields'' $X$ of the  
	probabilistic influence of variables not in $\blanket{X}$. Formally, for every $W
	\neq X \in \set{V}$, 
	\begin{equation} \label{eq:blanketDef}
		W \notin \blanket{X} \Rightarrow \ci{X}{W}{\blanket{X} - \{X, W\}} 
	\end{equation}
	A \emph{Markov boundary} is a minimal Markov blanket, \ie, non of its proper subsets
	satisfy Eq.(\ref{eq:blanketDef}). 
\end{definitionStar}

The substraction of $X$ and $Y$	from $\blanket{X}$ in
the consequent of Eq.~(\ref{eq:blanketDef}) is redundant, as neither $X$ nor $Y$ are in the blanket, and it is made
explicit for later convenience. Also, unless explicitly stated, from now on any mention of Markov 
blanket refers to a minimal Markov blanket, that is, to a boundary.

It can be proven that for Markov boundaries, the opposite direction of Eq.~(\ref{eq:blanketDef})
also holds, that is, 

\begin{lemma} \label{thm:blanketDefOpp}
	For every $W \neq X \in \set{V}$, if $\blanket{X}$ is the Markov boundary of $X$, then
	\begin{equation} \label{eq:blanketDefOpp}
			W \notin \blanket{X} \Leftarrow \ci{X}{W}{\blanket{X} - \{X, W\}}, 
	\end{equation}
\end{lemma}
The proof of this Lemma is discussed in Appendix \ref{app:app1}.

We can now present the Markov-blanket closure used by IBMAP-HC. We do it through the
following Theorem: 

\begin{theorem} [Markov-blanket closure] \label{thm:closure}
	Let $\set{V}$ be a domain of random variables, $G$ an independence structure over
	$\set{V}$, and let $\blanket{X}$ denote the Markov boundary of  
	a variable 	$X \in \set{V}$. Then, the set of independences 
	\begin{eqnarray} \label{eqn:MBclosure}
	  \closure_{MB}(G) = & & \Big\{~ \ci{X}{Y}{\blanket{X} - \{X,Y\} } \mbox{  \Large{$\mid$}  } X, Y\neq X \in \set{V}, ~ (X,Y) \notin E(G) ~ \Big\}~  \bigcup ~ \nonumber \\  
	  & & \Big\{  \cd{X}{Y}{\blanket{X} - \{X,Y\} } \mbox{  \Large{$\mid$}  } X, Y\neq X \in \set{V}, ~ (X,Y) \in E(G) \Big\} 
	\end{eqnarray}
	is a closure of $G$. That is, for each variable $X$ and each other variable $Y \neq X$,
	if the pair $(X,Y)$ is an edge in $G$ then add $\cd{X}{Y}{\blanket{X} - \{X,Y\}}$ to the
	closure, otherwise add $\ci{X}{Y}{\blanket{X} - \{X,Y\}}$. 
\end{theorem}

We need one final result to prove the Theorem, a relation between Markov boundaries and
independence structures:

\newtheorem*{corollaryTwo}{Corollary 2}
\begin{corollaryTwo}[\citet{pearl88}, p.98] 
The independence structure $G$ of any strictly positive distribution over $\set{V}$ can
be constructed by connecting each variable $X \in \set{V}$ to  
all members of its Markov boundary $\blanket{X}$. Formally,
	\begin{equation} \label{eq:blanketEdges}
		\forall Y \in \blanket{X} \Leftrightarrow (X,Y) \in E(G)
	\end{equation}
\end{corollaryTwo}

\begin{proofJMLR}[Theorem \ref{thm:closure}]

To prove that $\closure_{MB}(G)$ determines $G$, we must prove the fact that 
an edge between $X$ and $Y$ exists or not is determined by the independencies 
in $\closure_{MB}(G)$. We do it separately for existence and non-existence of edges.

\textbf{For edge existence:}

Let $(X,Y) \in E(G)$. Is this edge determined by $\closure_{MB}(G)$? By definition of $\closure_{MB}(G)$, 
the assertion $\cd{X}{Y}{{\blanket{X}-\{X,Y\}}}$ must be in $\closure_{MB}(G)$. It is
sufficient then to prove that $\cd{X}{Y}{{\blanket{X}-\{X,Y\}}} \Rightarrow (X,Y) \in
E(G)$. This follows from the counter positive of Eq.~(\ref{eq:blanketDef}) and
Eq.~(\ref{eq:blanketEdges}). 

\textbf{For edge absence:}
Let $(X,Y) \notin E(G)$. Is this lack of edge determined by $\closure_{MB}(G)$? By definition of $\closure_{MB}(G)$, 
the assertion $\ci{X}{Y}{{\blanket{X}-\{X,Y\}}}$ must be in $\closure_{MB}(G)$. It is
sufficient then to prove that $\ci{X}{Y}{{\blanket{X}-\{X,Y\}}} \Rightarrow (X,Y) \notin E(G)$.
This follows from Eq.~(\ref{eq:blanketDefOpp}) and Eq.~(\ref{eq:blanketEdges}).
\end{proofJMLR}

\subsubsection{Complexity of IBMAP-HC \label{sec:IBMAP-HC-complexity}}

Let's first discuss the computational complexity of IB-score using the Markov blanket
closure. According to Eq.~(\ref{eq:complexityIBScore}), this complexity is
$O(|\closure_{MB}|N\tau^*)$.  To obtain $|\closure_{MB}|$ note that for each variable $X$ in
$\set{V}$, and each other variable $Y \neq X$, there is 
either an edge or not, but both cases cannot happen. Thus, for any pair $X \neq Y$ either
the condition on the l.h.s. of the union in Eq.~(\ref{eqn:MBclosure}) is true, or the
condition in the r.h.s. is true, so only one independence assertion is added per pair of
variables. Therefore, $|\closure_{MB}| = n(n-1)$. We can also note that $\tau^* = 2 +
|\blanket{X^*}|$, where $X^* $ is the variable with the largest blanket.

A straight-forward implementation of the hill climbing computes the IB-Score for each of
the $n(n-1)/2$ neighbors, resulting in a total complexity of $O \left(
n^2(n-1)^2N\tau^*/2 \right)$. There is a trick, however, for reducing the complexity of each
neighbor's IB-Score by one order of magnitude. The difference of each neighbor with the
currently visited structure $G$ is exactly one edge, say $(X,Y)$. In the two cases of
an edge being removed, or an edge being added, the blanket of both $X$ and $Y$ would
change, and thus the conditionant of all $2(n-1)$ independence assertions in the closure
containing either $X$ or $Y$. All other independence assertions would remain exactly as
in $G$, and thus, an incremental computation of the IB-score would require only $2(n-1)$
statistical tests, resulting in a time complexity of the (incremental) IB-score of
$O \left( 2(n-1)N\tau^* \right)$, and a time complexity of one hill climbing step
of $O \left( n(n-1)^2N\tau^*  \right)$.  That is, a reduction of a factor $n/2$ with
respect to the non-incremental alternative.    

Finally, if we denote by $M$ the total number of hill climbing iterations, the overall
complexity of the IBMAP-HC algorithm is $O \left( Mn(n-1)^2N\tau^*  \right)$.
Unfortunately $M$ is unknown, and can only be obtained through empirical measurement. We
thus report it in our experiments.    

\subsection{IBMAP-TS: Independence-based MAP structure learning using tree search and
GSMN based closure.} 
\label{sec:ibmapts}

In this section we introduce another algorithm, \textbf{IBMAP-TS} (Independence-based
maximum a posteriori using tree search) for learning an independence structure using the
independence-based MAP approach of Section~\ref{sec:approach}. This algorithm implements
the maximization through a tree search, the \emph{uniform cost algorithm}, using an
\emph{algorithm-based closure} described briefly above in
Example~\ref{ex:algorithm-based-closure}. The resulting algorithm is not efficient,
requiring an exponential number of statistical tests to find the structure with maximum
IB-score. However, we believe its description here and the presentation of experimental
results later,  to be helpful in two ways. First, the closure presented is generic, in
the sense that can be easily instantiated for other independence-based algorithm besides
the GSMN algorithm used here. Also, although the algorithm is exponential in time, the
quality improvements obtained through experimentation help to reinforce the hypothesis
that the IBMAP approach improves the quality of the structures learned.         

Let's start formalizing the process followed by an independence-based algorithm. As
mentioned above in Section~\ref{sec:literature}, an independence-based algorithm proceeds
as follows: at each iteration $i$, it proposes a triplet $T_i$ and performs a statistical
independence test (SIT) on data to obtain an independence assertion $T_i=t_i^{SIT}$. The
superscript $SIT$ is included for later convenience, and denotes that the independence
value considered is the one indicated by the statistical independence test. Which triplet
is selected at iteration $i$ depends on $T_{1:i-1} = t_{1:i-1}^{SIT}$, that is, on the
sequence $T_{1:i-1}$ of triplets proposed so far (in iterations $1$ through $i-1$), and
the independence values $t_{1:i-1}^{SIT}$ that statistical tests assigned them (with
$T_{1:0}$, $t_{1:0}^{SIT}$ indicating the empty set of triplets and assertions,
respectively). Each iteration of an independence-based structure learning algorithm can
thus be summarized as follows:       
\begin{eqnarray}
	T_i  &\longleftarrow & A(T_{1:i-1} = t_{1:i-1}^{SIT}) \\
	t_i^{SIT} &\longleftarrow & SIT(T_i) \nonumber
\end{eqnarray}
where $A$ denotes the operator that decides the next triplet.

These algorithms proceed until the set of independence assertions done so far is
sufficient for determining a structure. By the definition of closure, this set is thus a
closure. We call the closure obtained in this way an \emph{algorithm-based closure}.    

How can we then use this closure in a MAP search? At each iteration $i$, we propose
considering, besides the independence assertion $T_i=t_i^{SIT}$, the alternative value
$T_i=\lnot t_i^{SIT}$. In other words, we propose to distrust the statistical test.
Interestingly, this change should not affect the algorithm. It would simply continue as
if the statistical test would have given the other value. Moreover, it would also find a
structure eventually, and the assertions found during its execution would thus be a
closure as well. Clearly, the structure found would be different. In summary, each such
bifurcation into $T_i=t_i^{SIT}$ and $T_i=\lnot t_i^{SIT}$ produces a different family
of structures whose posterior can be computed using as closure the assertions obtained
during its execution.        

To finalize, we can notice each bifurcation splits the sequence of independences in two,
each of which is recursively split in the next iteration. This can be modeled by a binary
tree, with each node corresponding to an independence assertion, and whose children
correspond to the two assignments for the triplet obtained by applying $A$ to the
assertions in the path from the root to the parent. An exception is the root node, which
represents a dummy empty sequence node. Each path from the root to a leave, not only
determines a structure, but it determines a closure $\{T_1=t_1, T_2=t_2, \ldots,
T_c=t_c\}$ for that structure. This closure can then be used to compute the IB-score of
the structure by multiplying the posterior of the independence assertion of each node in the
path, \ie, $\prod\limits_{i=1}^c \Pr(T_i =t_i\mid \data)$. Under this view, finding the
structure with maximum IB-score (\ie, the MAP structure) can be done through any tree
search algorithm using as successor function the operator $A$, and as cost of each action
$-\log{\Pr(T_i = t_i\mid \data)}$. Since the log of a product is the sum of the logs, and
the log is a monotonous function, finding the maximum IB-score $\prod\limits_{i=1}^c
\Pr(T_i = t_i\mid \data)$ is equivalent to finding the path with the minimal path-cost,
\ie, the sum of the costs of nodes in the path or $\sum\limits_{i=1}^c -log{\Pr(T_i =
t_i\mid \data)}$.            

We implemented and tested IBMAP-TS with the closure constructed using the GSMN algorithm
and uniform cost strategy to search the tree. We chose
GSMN to demonstrate our approach for being a well established independence-based Markov
networks structure learning algorithm. This algorithm uses the \textbf{GS} algorithm
\citep{MARGARITIS00} for learning the Markov blanket of every variable $X \in \set{V}$,
and uses Corollary $2$ to build the structure. Results comparing the
quality of structures learned by GSMN versus IBMAP-TS with uniform cost and GSMN based
closure are shown in the experimental results section. One important advantage of
uniform cost is that its solutions are optimal. To improve over its exponential runtime,
we tested a few heuristics (not reported here), but with no success in avoiding
exponential runtimes. We thus limit our findings to uniform cost.          

\begin{figure}[t]
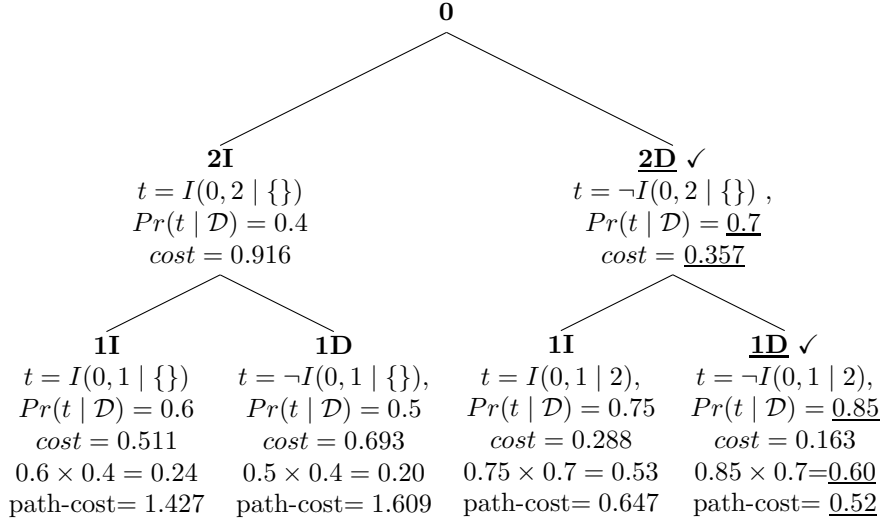

\small
\Tree 
	[.{
	  \textbf{0}
	  }
	  [.{
		\textbf{2I} \\
		$t=\ci{0}{2}{\{\}}$ \\ 
		$Pr(t\mid\mathcal{D})=0.4$ \\ 
		$cost=0.916$
	     }
	        {\textbf{1I} \\
			$t=\ci{0}{1}{\{\}}$ \\
			$Pr(t\mid\mathcal{D})=0.6$ \\ 
			$cost=0.511$ \\ 
			$0.6 \times 0.4=0.24$ \\
			path-cost$=1.427$
		  } 
		{\textbf{1D} \\
			$t=\cd{0}{1}{\{\}}$,\\
			$Pr(t\mid\mathcal{D})=0.5$ \\ 
			$cost= 0.693$ \\
			$0.5 \times 0.4=0.20$ \\
			path-cost$ = 1.609$
		} 
	   ] 
	   [.{
		\underline{\textbf{2D}} \checkmark \\
			$t=\cd{0}{2}{\{\}}$ ,\\
			$Pr(t\mid\mathcal{D})=\underline{0.7}$ \\ 
			$cost=$ \underline{$0.357$}
	     }
		{\textbf{1I} \\
			$t=\ci{0}{1}{2}$,\\
			$Pr(t\mid\mathcal{D})=0.75$ \\ 
			$cost = 0.288$ \\
			$0.75 \times 0.7=0.53$ \\
			path-cost$=0.647$
			} 
		{\underline{\textbf{1D}} \checkmark \\
			$t=\cd{0}{1}{2}$,\\
			$Pr(t\mid\mathcal{D})=\underline{0.85}$ \\ 
			$cost = 0.163$ \\
			$0.85 \times 0.7$=\underline{$0.60$} \\
			path-cost= \underline{$0.52$}
			}
	    ]
	 ]
\caption{\label{fig:exampletree}Example partial binary tree expanded by IBMAP-TS with
uniform cost search and GSMN-based closure.} 
\end{figure}

Fig.(\ref{fig:exampletree}) shows an example (partial) search tree for a small system of
three random variables $\set{V}=\{0,1,2\}$. Each node is annotated with both the triplet
and independence assertion of the triplet (\eg, node 1I is labeled by
$\ci{0}{1}{\{\}}$), the posterior probability of this independence assertion
(e.g.,$\Pr(\ci{0}{1}{\{\}} \mid \data) = 0.6$), the local cost (e.g.,$-
\log(\Pr(\ci{0}{1}{\{\}} \mid \data)) = 0.511$), the partial computation of the IB-score
(\eg, for $1I$ this would $0.4 \times 0.6$, the product of costs of all nodes from the
root to $1I$), and finally, the partial path-cost as the minus log of the product (e.g.,
$1.427$ for $1I$).  The node with the checkmark (and underlined) is the one with lowest
path-cost, and thus the next in line to be expanded by uniform-cost.

\section{Experimental evaluation}\label{sec:experiments}

We describe several experiments for testing the effectiveness of the IB-score for
improving the quality of independence structures discovered by our novel
independence-based algorithms IBMAP-HC and IBMAP-TS (c.f. \S\ref{sec:algorithms}). The
experiments also corroborate the practicality of IBMAP-HC, \ie, its polynomial running
time, as discussed theoretically in Section \ref{sec:IBMAP-HC-complexity}.

\subsection{Experimental setup}\label{sec:setup}
We discuss here the performance measures we use for testing the quality of output
networks, and the time complexity of the algorithms, as well the datasets on which these
performance was measured. 

\subsubsection{Datasets}

We ran our experiments on benchmark (real-world) and sampled (artificial) datasets.
Real world datasets allow an assessment of the performance of our algorithms in
realistic settings with the disadvantage of lacking the solution network, thus resulting
in approximate measures of quality. We used the publicly
available benchmark datasets obtained from the UCI Repositories of machine learning
\citep{Asuncion+Newman:2007} and KDD datasets \citep{Hettich+Bay:1999}.               
Artificial datasets, although more limited
in the scope of the results, are sampled from known networks
allowing a more systematic and controlled study of the performance of our algorithms.

Using a Gibbs sampler, data was sampled from several randomly generated undirected
graphical models, each with a randomly generated graph and parameters.  We considered
models with different number of variables $n$ (\eg, 
$n=12,50$), and different number $\tau$ of 
neighbors per node (we used $\tau=1,2,4,8$ in all our experiments). 
For a given $n$ and $\tau$, $10$ graphs were generated, connecting randomly and
uniformly, each of the $n$ nodes to $\tau$ other nodes. This was achieved by connecting
each node $i$ with the first $\tau$ nodes of a random permutation of
$[1,\ldots,i-1,i+1,\ldots,n]$. For each of the generated graphs, a set of parameters
was generated randomly following a 
procedure described in detail in \cite{bromberg&margaritis09b} that guarantees
dependencies remains strong for variables distant in the network. Finally, one dataset
was sampled for each of the generated pair of graph and set of parameters. 

\subsubsection{Computational cost measure\label{sec:complexity}}

One of the hypothesis we wanted to prove with our experiments is the polynomial runtime
of IBMAP-HC. In Section \ref{sec:IBMAP-HC-complexity} we found the time complexity of
IBMAP-HC to be $O \left( Mn(n-1)^2 N \tau^* \right)$. For this expression to be
polynomial both $M$ and $\tau^*$ should at most grow polynomially with $n$. The size of
the largest triplet test $\tau^*$ depends only in the connectivity of the networks, which
we kept fixed at $\tau$. Instead, we have no elements to predict the behavior of $M$, as
it depends solely on the landscape of the score function. We thus report $M$ in our
experiments as a measure of complexity.

\subsubsection{Error measures}
We measured quality through two types of errors between the output network and a model considered
to be the correct one: the \emph{edges} and \emph{independences
Hamming distance}. For the case of artificial datasets the comparison was done against
the true, known, underlying network, whereas for real datasets the comparison was done
against the data itself. Let us discuss these quantities in detail:

	\begin{itemize}
	\item 
        The \textit{edge Hamming distance} $H_{\textsc{E}}(G,G')$ between two graphs $G$
		and $G'$ of equal number of nodes, represents the number of edges that
		exist in $G$, and do not exist in $G'$, and vice versa. Put another way, it measures
		the minimum number of edge substitutions required to change one graph into the other.
		To measure the error of a structure $G$ output by a structure learning
		algorithm we measure its edges Hamming distance $H_{\textsc{E}}(\gstar, G)$, or simply
		$\He{}$, with the solution network $\gstar$. Formally, if we define the following
		indicator function for evaluating the existence of an edge between two variables $X$
		and $Y$ in $G$,          
	\begin{equation}
                \textsc{E}^{G}(X,Y) = \left\{
                \begin{array}{c l}
                1 & (X,Y) \mbox{ is an edge in } G \\
                0 & \mbox{otherwise.}
                \end{array}
                \right.
        \end{equation}
 
	the edges Hamming distance is defined as

         \begin{equation} \label{eq:he}
		   \He{} = \Bigg{\vert} \Bigg{\{} X,Y \in \set{V} , X \neq Y ~\bigg{\vert}~
		   \textsc{E}^{G}(X,Y) \neq \textsc{E}^{G^\star}(X,Y) \Bigg{\}} \Bigg{\vert} 
         \end{equation}

	\item
	Given two probability distributions $P$ and $P'$ over the same set of variables
	$\set{V}$, we define the \textit{independences Hamming distance} $H_{\textsc{I}}(P,P')$
	between them to be the (normalized) number of matches in a comparison of the
	independence assertions
	that holds in $P$ and $P'$. That is,
	if $\mathcal{T}$ denotes the set of all possible triplets over $\set{V}$,   
	it is checked for how many triplets $t \in \mathcal{T}$, $t$ is independent (or
	dependent) in both distributions, and then normalized by $\mathcal{T}$.
	Unfortunately, the size of $\mathcal{T}$ is exponential. We thus compute the
	approximate Hamming distance $\hat{H}_{\textsc{I}}(P,P')$ over a randomly sampled subset
    of $\widehat{\mathcal{T}}$, uniformly distributed for each conditioning set cardinality.
	In all our experiments we used $| \widehat{\mathcal{T}}| = 2,000$, constructed as follows:
	for each $m = 0,1,2,\ldots, n-2$, $|\widehat{\mathcal{T}}|/(n-1)$ triplets
	$\triplet{X}{Y}{\set{Z}}$ with cardinality $|\set{Z}|=m$ were sampled randomly and
	uniformly by first generating a random permutation $[\pi_1, \pi_2, \ldots, \pi_{n}]$
	of the set of all variables $\set{V}$, and the assigning $X = \pi_1$, $Y = \pi_2$,
	and $\set{Z} = [\pi_3, \ldots, \pi_m]$. 
	   
	In what follows we are not given a fully specified distribution. Instead,
	dependending on the type of dataset being artificial or real, we are given
	the independence structure or a sample (\ie, a dataset), respectively.
	Independencies, however, can be measured over both independence structures and
	datasets, so in both cases we can measure (approximately) the independence Hamming
	distance. Let's consider both cases.
	
	To estimate the error of structures output by our algorithms when run over
	artificial datasets, we measured their independence Hamming distance
	$H_{\textsc{I}}(G,\gstar)$ (or simply $\Hitrue{}$), with the
	underlying true network $\gstar$, querying independences directly on the structures
	using vertex-separation. Formally, if we denote $I^{G^\star}(t)$ the result of a test $t \in
	\widehat{\mathcal{T}}$ performed on the true model, and by $I^{G}(t)$ the result of the same test
	$t$ performed on a model $G$, then the independence Hamming distance $\Hitrue{}$ is
	defined formally as: 	 
	
	\begin{equation} \label{eq:hinet}
		\Hitrue{} = \frac{1}{\mid \widehat{\mathcal{T}}\mid}~\Bigg{\vert}\{ t\in
		\widehat{\mathcal{T}} ~\bigg{\vert}~ I^{G}(t)\neq I^{G^\star}(t)\}\Bigg{\vert}
	\end{equation}
	
	In experiments over real datasets the underlying structure $\gstar$ is unknown. We
	thus conducted experiments learning the structure over smaller datasets with sizes $1/3$ and
	$1/5$ of the input dataset $\mathcal{D}$, and compared the independencies in
	the output structure $G$, and the complete dataset $\mathcal{D}$. The resulting
	Hamming distance is denoted $\widehat{H}_{\textsc{I}}(G,\mathcal{D})$, or simply
	$\Hidata{}$.  Formally, if we denote by $I^{\mathcal{D}}(t)$ the result of a test $t \in
	\widehat{\mathcal{T}}$ performed on the complete dataset, then $\Hidata{}$ is defined as:      

	\begin{equation} \label{eq:hidata}
		\Hidata{} = \frac{1}{\mid\widehat{\mathcal{T}}\mid}~\Bigg{\vert}\{
		t\in\widehat{\mathcal{T}} ~\bigg{\vert}~ I^{G}(t)\neq I^{\mathcal{D}}(t)\}\Bigg{\vert}
	\end{equation}
	
	\end{itemize}

We are interested in comparing the errors of structures output by our algorithms
(IBMAP-HC , IBMAP-TS) and GSMN, the competitor. For that we report the
ratio $r =\frac{H(G_{HC})}{H(G_{GSMN})}$ of the errors of the network $G_{HC}$ output by
IBMAP-HC against the error of the network $G_{GSMN}$ output by GSMN. Similarly, for
network $G_{TS}$ output by IBMAP-TS, we report the ratio $r =\frac{H(G_{TS})}{H(G_{GSMN})}$.
Being three different types of errors, the edge Hamming distance $\He{}$ and the two
independence Hamming distances $\Hitrue{}$ and $\Hidata{}$, this results in six possible
ratios, three for $HC$ and three for $TS$:

   \begin{tabular}[m]{ccccccccc} 
	 \\
	$\rHe{HC}$     &=& $\frac{\He{HC}}{\He{GSMN}}$,         & 
 	$\rHitrue{HC}$ &=& $\frac{\Hitrue{HC}}{\Hitrue{GSMN}}$, &
 	$\rHidata{HC}$ &=& $\frac{\Hidata{HC}}{\Hidata{GSMN}}$ \\
 	$\rHe{TS}$     &=& $\frac{\He{TS}}{\He{GSMN}}$,         &
 	$\rHitrue{TS}$ &=& $\frac{\Hitrue{TS}}{\Hitrue{GSMN}}$, &
 	$\rHidata{TS}$ &=& $\frac{\Hidata{TS}}{\Hidata{GSMN}}$ \\
    \end{tabular}

These ratios allows a quick comparison between the two algorithms it involves as a ratio
equal to one means the same error in the structures they output, a ratio lower than
one means a reduction in the error of the structures output by our algorithms (HC or TS)
and a ratio greater than one means a reduction in quality by our algorithms.

\subsection{Experimental results\label{sec:experimentsresults}}

We show now the results of our experiments.


\begin{figure}
   \centering
   \begin{tabular}[m]{c||c|c} 
	\multicolumn{3}{c}{Comparison of Hamming distances, $n=12$}\\ \hline
	 & & \\ 
 	 $\tau$ & $r^{\star}_{\mathtt{E}}$ & $r^{\star}_{\mathtt{I}}$ \\ \hline
	& & \\ 
 	\large{\textbf{1}}  &
	\epsfig{figure=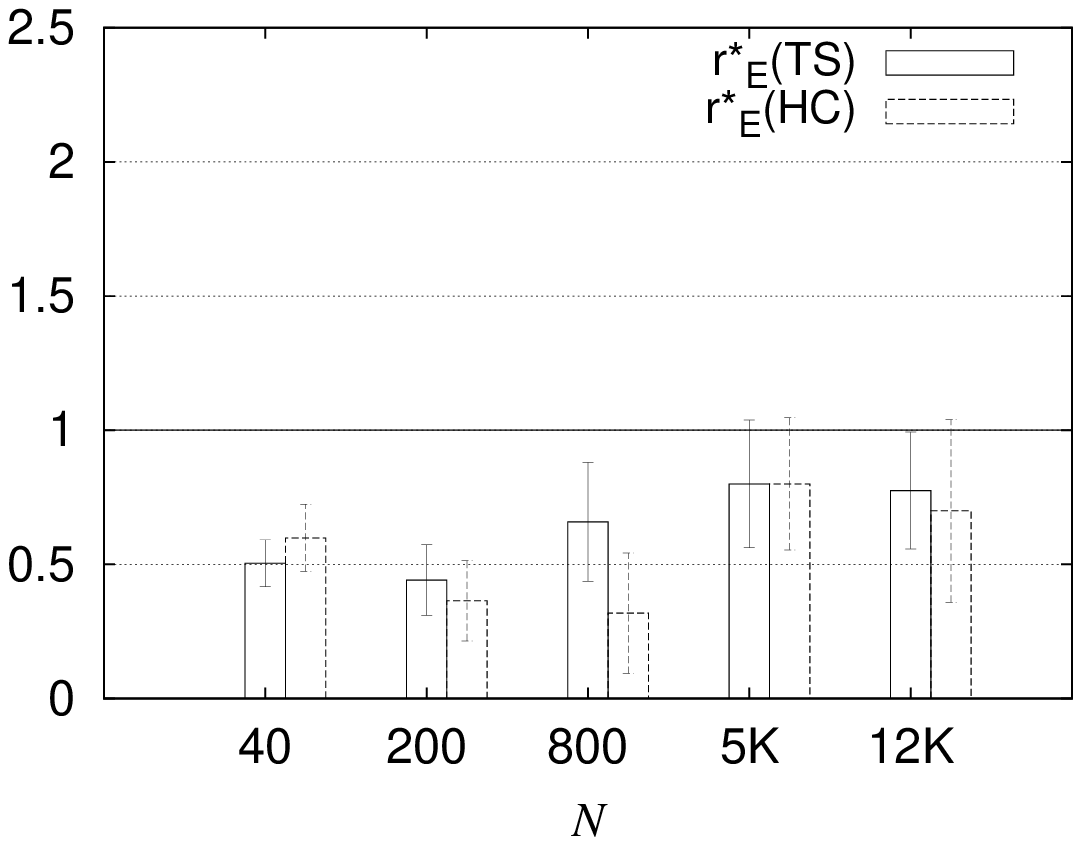,width=5.5cm} & 
	\epsfig{figure=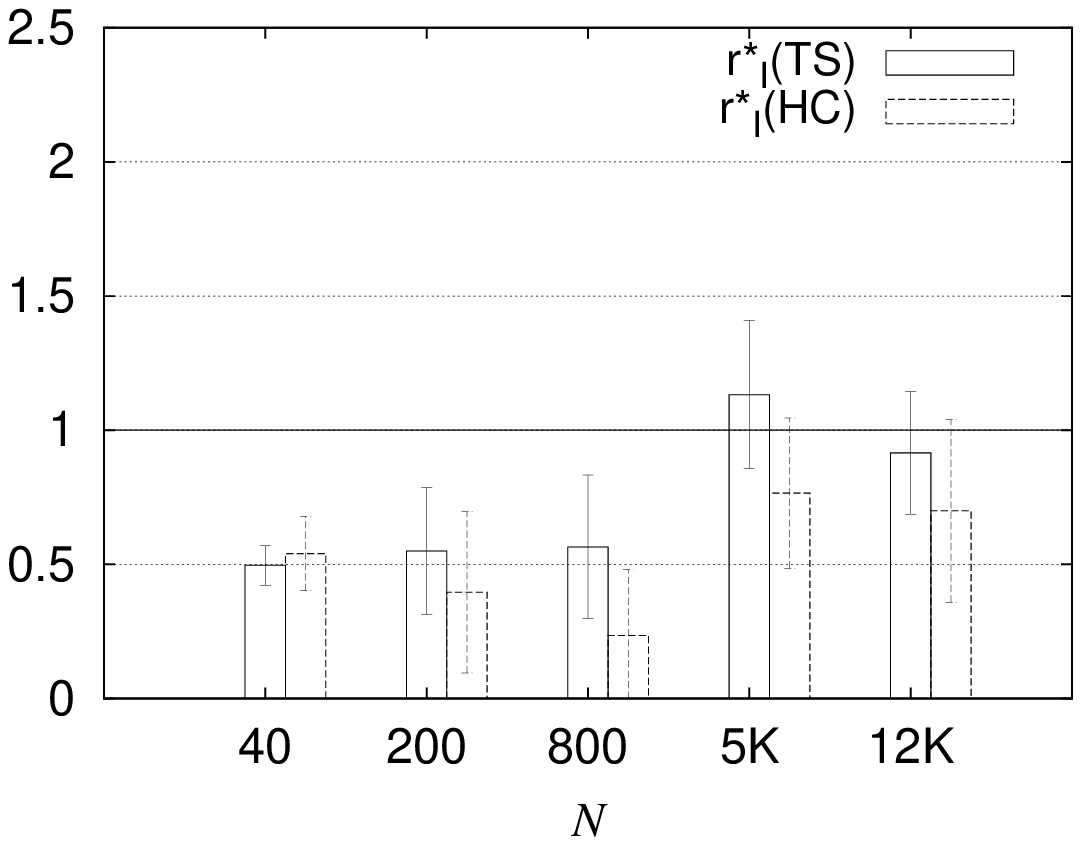,width=5.5cm} \\ \hline
	& & \\ 
	\large{\textbf{2}}  &
	\epsfig{figure=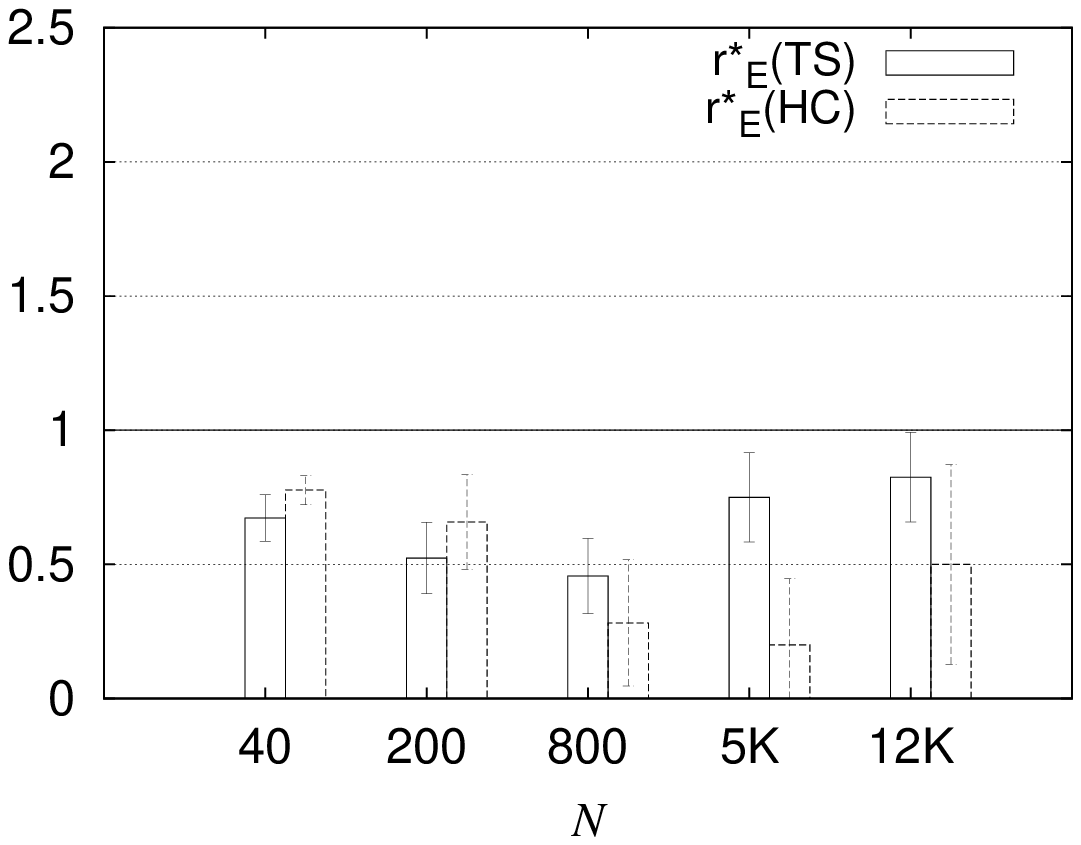,width=5.5cm} &
 	\epsfig{figure=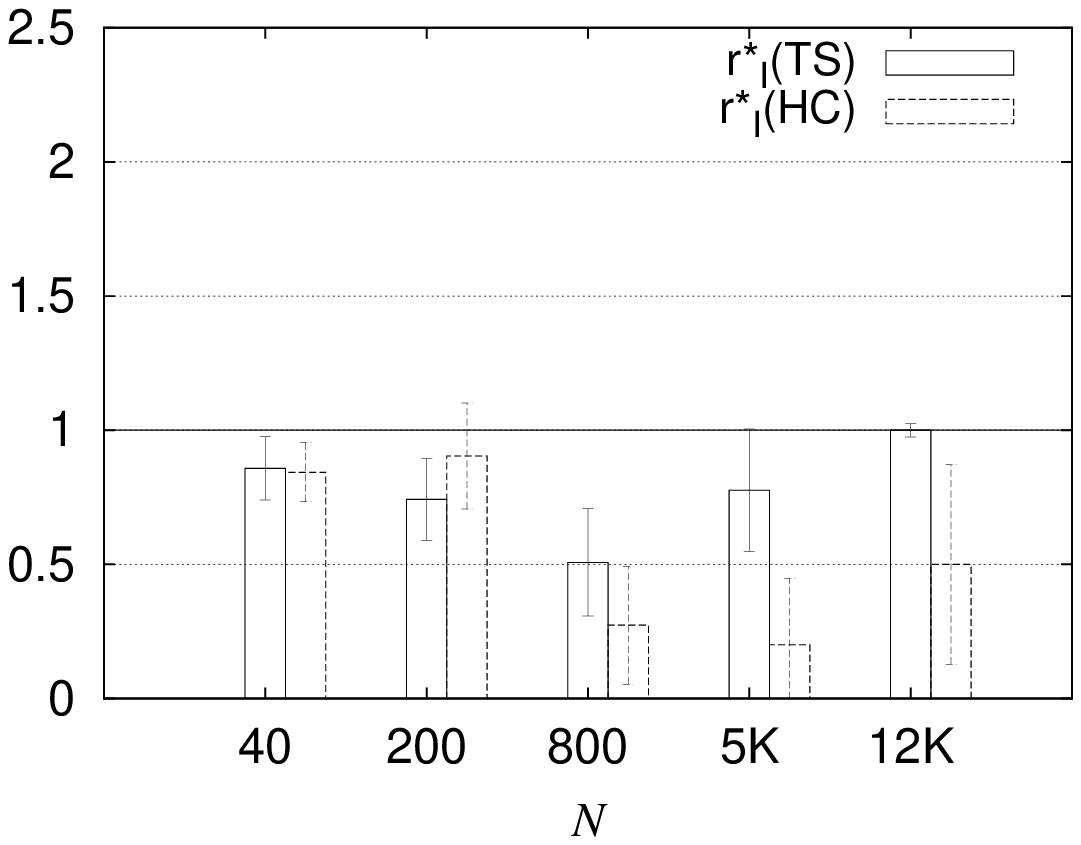,width=5.5cm} \\ \hline
	& & \\ 
 	\large{\textbf{4}} &
	\epsfig{figure=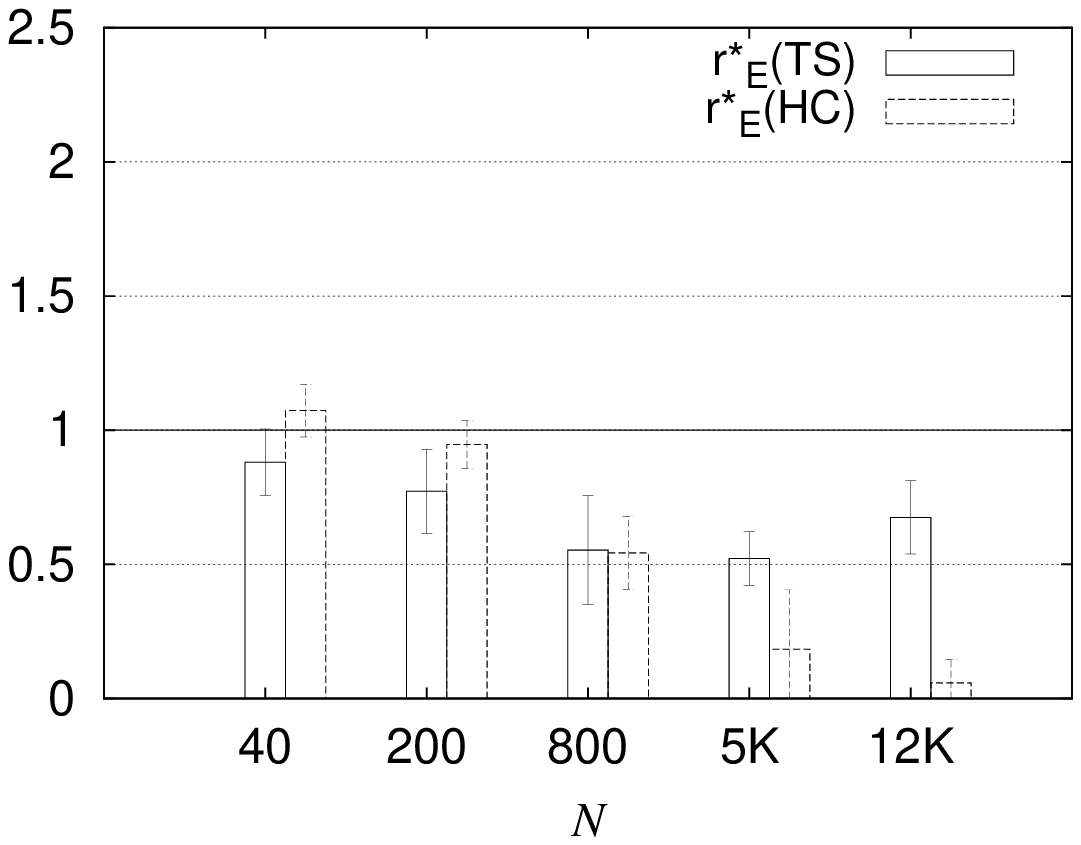,width=5.5cm} & 
	\epsfig{figure=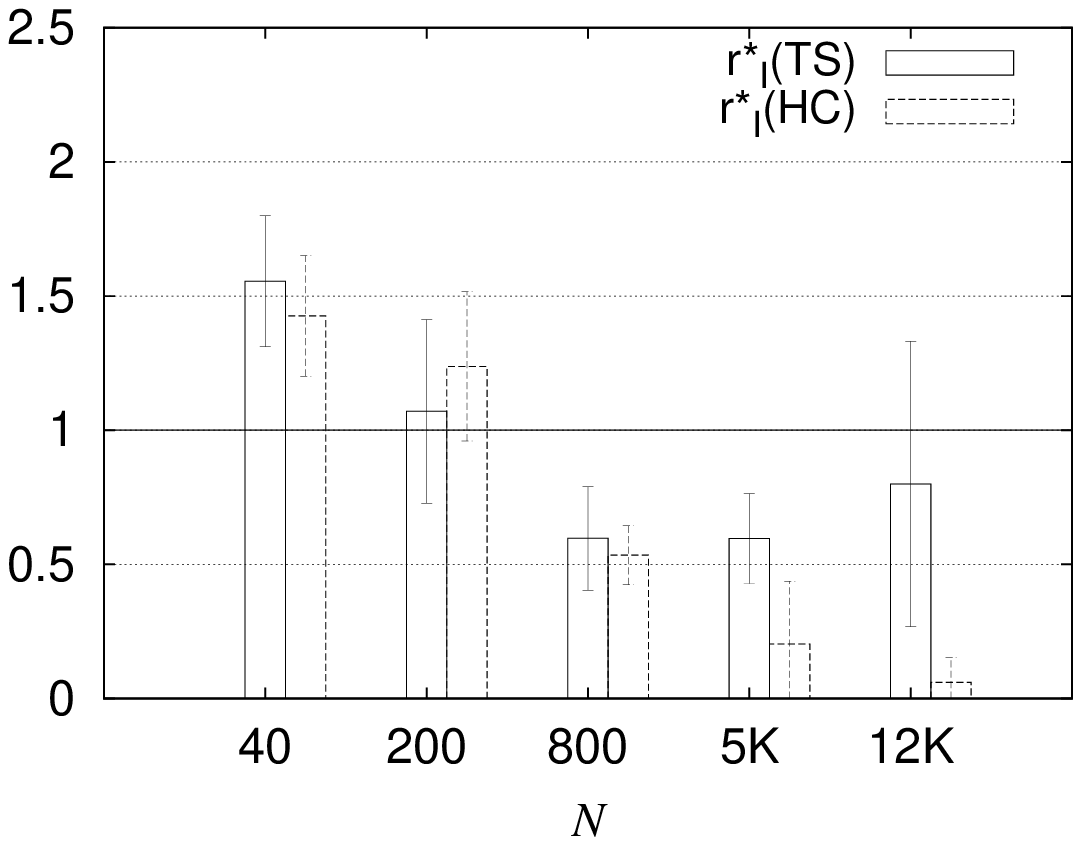,width=5.5cm} \\ \hline
	& & \\ 
	\large{\textbf{8}}  &
	\epsfig{figure=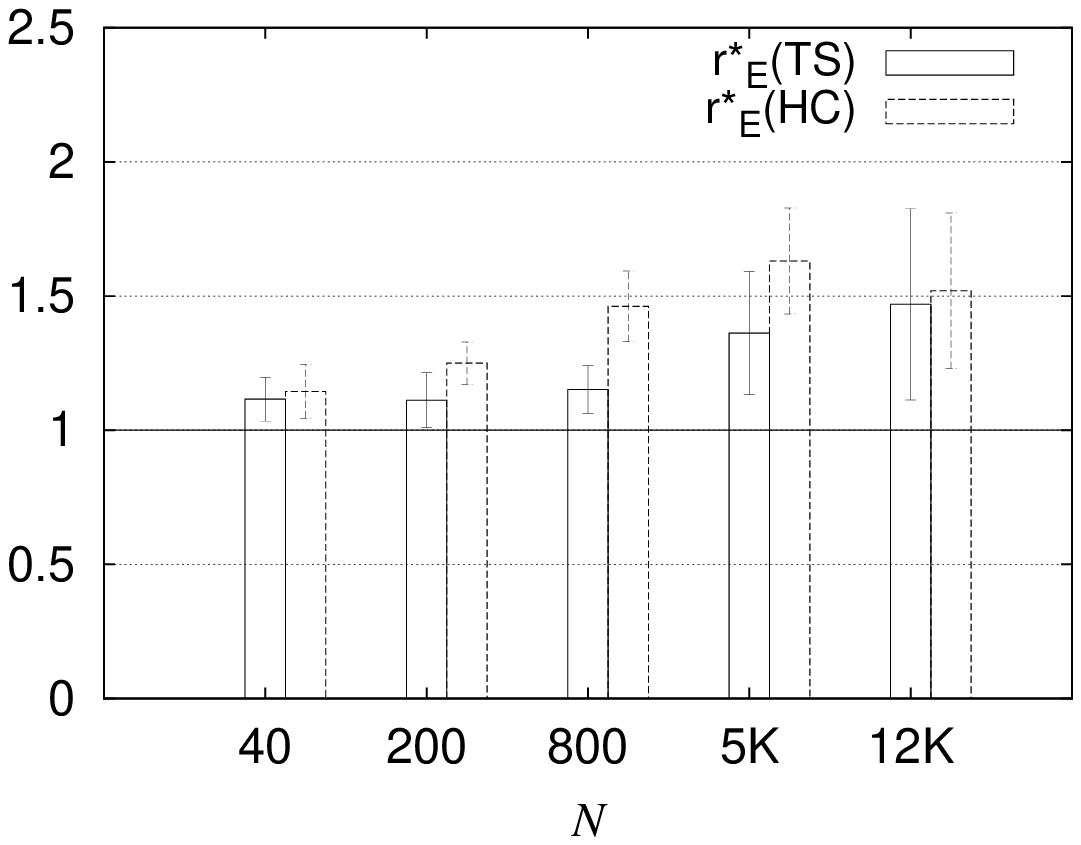,width=5.5cm} & 
	\epsfig{figure=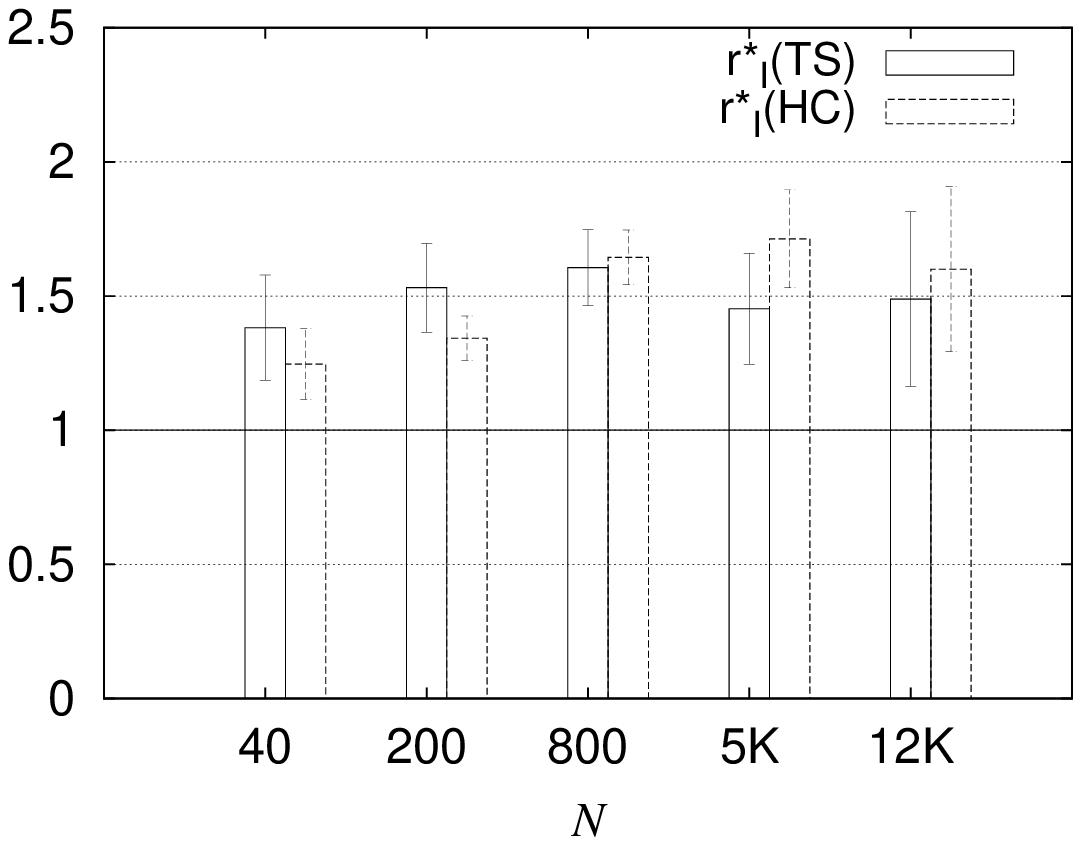,width=5.5cm} \\ \hline
    \end{tabular}
   \caption{\label{fig:n12He}  Average of $\rHe{TS}$, $\rHe{HC}$ (left column), $\rHitrue{TS}$ and
   $\rHitrue{HC}$ (right column) over $10$ datasets with $n=12$ and $\tau=1,2,4,8$ (rows). Error bars
   represents standard deviation. The smaller the value, more important is the
   improvement, with values greater (smaller) than 1 represent reduction
   (increase) in the quality. }    
 \end{figure}

\subsubsection{Sampled Data Experiments \label{sec:sampledexperiments}}

In our first set of experiments, we demonstrate that our two IBMAP algorithms IBMAP-HC and
IBMAP-TS are successful in improving the quality over artificially generated datasets by
comparing their edge errors $\He{HC}$ and $\He{TS}$ as well as their independence errors
$\Hitrue{HC}$ and $\Hitrue{TS}$ against the corresponding errors $\He{GSMN}$ $\Hitrue{GSMN}$
of GSMN.

As the IBMAP-TS algorithm is impractical for larger domains, we considered two
scenarios, one over underlying models of size $n=12$, comparing errors of both IBMAP-HC and
IBMAP-TS against GSMN, and one over larger models of size $n=50$, comparing only
IBMAP-HC against GSMN (with results shown in ). To assess our algorithms over different conditions of
reliability and connectivity, we ran them over datasets with increasing number of
datapoints $N=40,200,800,5000,12000$, sampled from networks of increasing node degrees
$\tau=1,2,4,8$. To increase statistical significance, $10$ datasets were sampled for 
each pair $(n, \tau)$, and for each of them, one subsample of size $N$ was
obtained by randomly selecting $N$ datapoints. 

Table \ref{table:n12He} shows the edge errors for $n=12$, reporting the mean values and
standard deviations (in parenthesis)  
of the edge errors $\He{GSMN}$, $\He{TS}$, and $\He{HC}$ of GSMN, IBMAP-TS and IBMAP-HC,
respectively, for the different conditions of connectivity and reliability. The last two
columns show their corresponding ratios $\rHe{TS}$ and $\rHe{HC}$, indicating in bold the
statistically significant improvements (i.e. ratios lower than $1$), and underlined the
statistically significant reductions in errors (\ie, ratios greater than $1$). These
error ratios are plotted also in Figure \ref{fig:n12He} (left column). The independence
errors $\Hitrue{GSMN}$, $\Hitrue{TS}$ and $\Hitrue{HC}$ for $n=12$ are shown in Table
\ref{table:n12Hi} with a similar structure, with the ratios also plotted in Figure 
\ref{fig:n12He} (right column).

\begin{table} [t]
\centering
\begin{scriptsize}
\begin{tabular}{|c|c|ccc||c|c|}
\hline
\multicolumn{7}{|c|}{ } \\ 
\multicolumn{7}{|c|}{Edges Hamming distances, $n=12$} \\ 
\hline
& & & & & & \\
$\tau$ & $N$ & $\He{GSMN}$ & $\He{TS}$ & $\He{HC}$ & $\rHe{TS}$ & $\rHe{HC}$  \\
& & & & & & \\
\hline
& & & & & & \\
\multirow{5}{*}{1}& 40 & 20.500(2.471) & 10.200(1.875) & 12.700(3.920) & \textbf{0.504(0.087)} & \textbf{0.599(0.125)} \\ 
 & 200 & 10.800(2.505) & 4.700(1.630) & 3.700(1.664) & \textbf{0.441(0.132)} & \textbf{0.364(0.150)} \\ 
 & 800 & 3.600(1.915) & 1.800(0.648) & 1.000(0.576) & \textbf{0.592(0.266)} & \textbf{0.318(0.225)} \\ 
 & 5000 & 0.800(0.928) & 0.900(1.125) & 0.400(0.493) & 1.025(0.056) & 0.800(0.247) \\ 
 & 12000 & 0.300(0.341) & 0.200(0.297) & 0.000(0.000) & 0.900(0.223) & 0.700(0.341) \\ 
& & & & & & \\
\hline
& & & & & & \\
\multirow{5}{*}{2} & 40 & 20.900(2.453) & 13.900(1.864) & 16.200(2.044) & \textbf{0.673(0.087)} & \textbf{0.777(0.055)} \\ 
 & 200 & 12.000(3.487) & 5.800(1.363) & 7.800(3.654) & \textbf{0.524(0.133)} & \textbf{0.658(0.177)} \\ 
 & 800 & 4.600(0.681) & 2.100(0.966) & 1.200(0.928) & \textbf{0.428(0.171)} & \textbf{0.282(0.235) }\\ 
 & 5000 & 2.100(1.022) & 1.400(0.681) & 0.300(0.476) & \textbf{0.725(0.242)} & \textbf{0.200(0.247) }\\ 
 & 12000 & 0.700(0.581) & 0.700(0.581) & 0.000(0.000) & 1.000(0.000) & \textbf{0.500(0.372) }\\ 
& & & & & & \\
\hline
& & & & & & \\
\multirow{5}{*}{4} & 40 & 21.000(3.308) & 18.100(2.408) & 22.200(2.717) & \textbf{0.880(0.124)} & 1.073(0.098) \\ 
 & 200 & 13.700(2.352) & 10.400(2.402) & 13.000(2.955) & \textbf{0.772(0.157)} & 0.947(0.088) \\ 
 & 800 & 8.300(1.332) & 4.400(1.669) & 4.600(1.457) & \textbf{0.553(0.203)} & \textbf{0.543(0.136)} \\ 
 & 5000 & 4.400(1.734) & 2.200(0.986) & 0.500(0.499) & \textbf{0.542(0.135)} & \textbf{0.183(0.221)} \\ 
 & 12000 & 2.400(0.828) & 1.500(0.685) & 0.200(0.297) & \textbf{0.575(0.245)} & \textbf{0.058(0.088)} \\ 
& & & & & & \\
\hline
& & & & & & \\
\multirow{5}{*}{8} & 40 & 30.600(4.475) & 33.600(3.261) & 34.300(2.702) & \underline{1.116(0.082)} & \underline{1.145(0.100)} \\ 
 & 200 & 23.200(2.634) & 25.600(3.086) & 28.900(3.414) & \underline{1.112(0.103)} & \underline{1.250(0.079)} \\ 
 & 800 & 17.400(2.402) & 19.900(2.648) & 25.000(2.532) & \underline{1.152(0.089)} & \underline{1.462(0.132)} \\ 
 & 5000 & 10.300(1.489) & 13.600(1.494) & 16.400(1.601) & \underline{1.362(0.230)} & \underline{1.631(0.198)} \\ 
 & 12000 & 9.200(2.913) & 12.200(2.613) & 12.900(2.810) & \underline{1.469(0.356)} & \underline{1.520(0.290) }\\ 
& & & & & & \\
\hline
\end{tabular}
\caption{\label{table:n12He}Average and standard deviation of $\He{GSMN}$, $\He{TS}$ and
$\He{HC}$ (columns) for datasets sampled from $10$ random networks of $ n = 12
$ and $\tau=1,2,4,8$ (rows). Last two columns shows $\rHe{TS}$ and $\rHe{HC}$, with quality
improvements when $r<1$ (in bold), and quality reductions when $r>1$ (underlined). }    

\end{scriptsize}
\end{table}

These results show that in most cases our proposed algorithms
present quality improvements over GSMN. For datasets with connectivities $\tau=1,2,4$,
the Hamming distances are better or 
equal for IBMAP-TS and IBMAP-HC on all cases in Table \ref{table:n12He}, and almost all
cases in Table \ref{table:n12Hi}. In some cases the improvement are drastic (e.g.
$\rHe{HC}=0.058(0.088)$, for $\tau=4$, $N=12000$ meaning that for approximately $6$ wrong
edges in $G_{HC}$ there are $100$ wrong edges in $G_{GSMN}$).
The pour results for datasets with connectivity $\tau=8$ can be explained by recalling
the approximation made in Eq.(\ref{eq:independenceApproximation}), namely, that
conditional independence assertions 
are mutually independent. This independence is not expected to hold over assertions
involving the common variables, with the stronger dependence for larger overlaps.

\begin{table}[h]
\centering
\begin{scriptsize}

\begin{tabular}{|c|c|ccc||c|c|}
\hline
\multicolumn{7}{|c|}{ } \\ 
\multicolumn{7}{|c|}{Independences Hamming distances, $n=12$} \\ 
\hline
& & & & & & \\
$\tau$ & $N$ & $\Hitrue{GSMN}$ & $\Hitrue{TS}$ & $\Hitrue{HC}$  & $\rHitrue{TS}$ & $\rHitrue{HC}$  \\
& & & & & & \\
\hline

& & & & & & \\
\multirow{5}{*}{1} & 40 & 0.593(0.058) & 0.293(0.048) & 0.322(0.098) & \textbf{0.497(0.074)} & \textbf{0.541(0.139)} \\ 
 & 200 & 0.397(0.093) & 0.186(0.069) & 0.113(0.043) & \textbf{ 0.550(0.236)} & \textbf{0.396(0.301)} \\ 
 & 800 & 0.164(0.096) & 0.084(0.054) & 0.020(0.013) & \textbf{0.565(0.267)} & \textbf{0.235(0.247)} \\ 
 & 5000 & 0.027(0.032) & 0.036(0.044) & 0.012(0.015)  & 1.132(0.276) & 0.766(0.280) \\ 
 & 12000 & 0.005(0.006) & 0.004(0.005) & 0.000(0.000) & 0.916(0.229) & 0.700(0.341) \\ 
& & & & & & \\
\hline
& & & & & & \\
\multirow{5}{*}{2} & 40 & 0.444(0.031) & 0.379(0.052) & 0.376(0.058) & \textbf{0.858(0.118)} & \textbf{0.843(0.111}) \\ 
 & 200 & 0.297(0.078) & 0.208(0.045) & 0.254(0.066) & \textbf{0.742(0.153)} & \underline{0.904(0.197)} \\ 
 & 800 & 0.166(0.040) & 0.094(0.050) & 0.045(0.036) & \textbf{0.507(0.199)} & \textbf{0.273(0.220)} \\ 
 & 5000 & 0.085(0.054) & 0.062(0.042) & 0.006(0.012) & \textbf{0.777(0.228)} & \textbf{0.200(0.247)} \\ 
 & 12000 & 0.030(0.032) & 0.031(0.033) & 0.000(0.000) & 1.000(0.025) & \textbf{0.500(0.372)} \\ 
& & & & & & \\
\hline
& & & & & & \\
\multirow{5}{*}{4} & 40 & 0.197(0.032) & 0.296(0.034) & 0.272(0.033) & \underline{1.555(0.244)} & \underline{1.426(0.225)} \\ 
 & 200 & 0.127(0.020) & 0.134(0.045) & 0.152(0.025) & 1.071(0.343) & 1.238(0.278) \\ 
 & 800 & 0.094(0.018) & 0.056(0.022) & 0.052(0.017)  & \textbf{0.597(0.193)} & \textbf{0.534(0.110)} \\ 
 & 5000 &  0.051(0.023) & 0.026(0.011) & 0.007(0.009) & \textbf{0.596(0.168)} & \textbf{0.203(0.234)} \\ 
 & 12000 & 0.027(0.011) & 0.022(0.012) & 0.002(0.003) & 0.800(0.532) & \textbf{0.060(0.093)} \\ 
& & & & & & \\
\hline
& & & & & & \\
\multirow{5}{*}{8} & 40 & 0.238(0.145) & 0.291(0.128) & 0.271(0.132) & \underline{1.382(0.196)} & \underline{1.246(0.132)} \\ 
 & 200 & 0.094(0.013) & 0.142(0.018) & 0.125(0.016) & \underline{1.531(0.166)} & \underline{1.343(0.083)} \\ 
 & 800 & 0.059(0.006) & 0.095(0.012) & 0.097(0.010)  &\underline{ 1.606(0.142) }& \underline{1.644(0.102)} \\ 
 & 5000 & 0.031(0.003) & 0.044(0.005) & 0.052(0.004) & \underline{1.453(0.207)} &\underline{ 1.713(0.182)} \\ 
 & 12000 & 0.024(0.007) & 0.033(0.008) & 0.036(0.008) & \underline{1.488(0.326)} & \underline{1.600(0.307)} \\ 
& & & & & & \\
\hline
\end{tabular}
\caption{\label{table:n12Hi}Average and standard deviation of $\Hitrue{GSMN}$,
$\Hitrue{TS}$ and $\Hitrue{HC}$ (columns) for datasets with $ n = 12 $ and $\tau=1,2,4,8$ (rows). Last
two columns shows $\rHitrue{TS}$ and $\rHitrue{HC}$, with quality improvements
when $r<1$ (in bold), and quality reductions when $r>1$ (underlined). }    
\end{scriptsize}
\end{table}

\begin{figure}
   \centering
   \begin{tabular}{c|c} 
	\multicolumn{2}{c}{Comparison of $\Hidata{HC}$, $n=50$}\\ \hline
 	$r^{\star}_{\mathtt{E}}$ & $r^{\star}_{\mathtt{I}}$ \\ 
 	\epsfig{figure=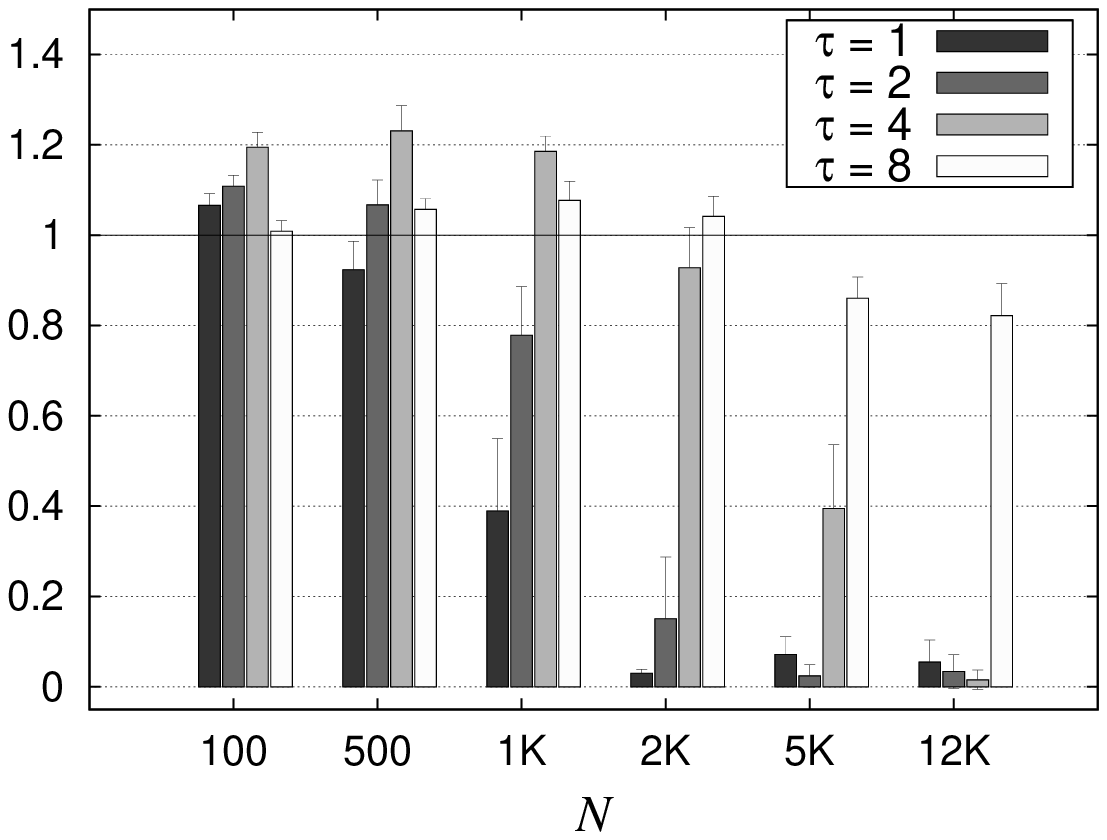,width=6cm} & \epsfig{figure=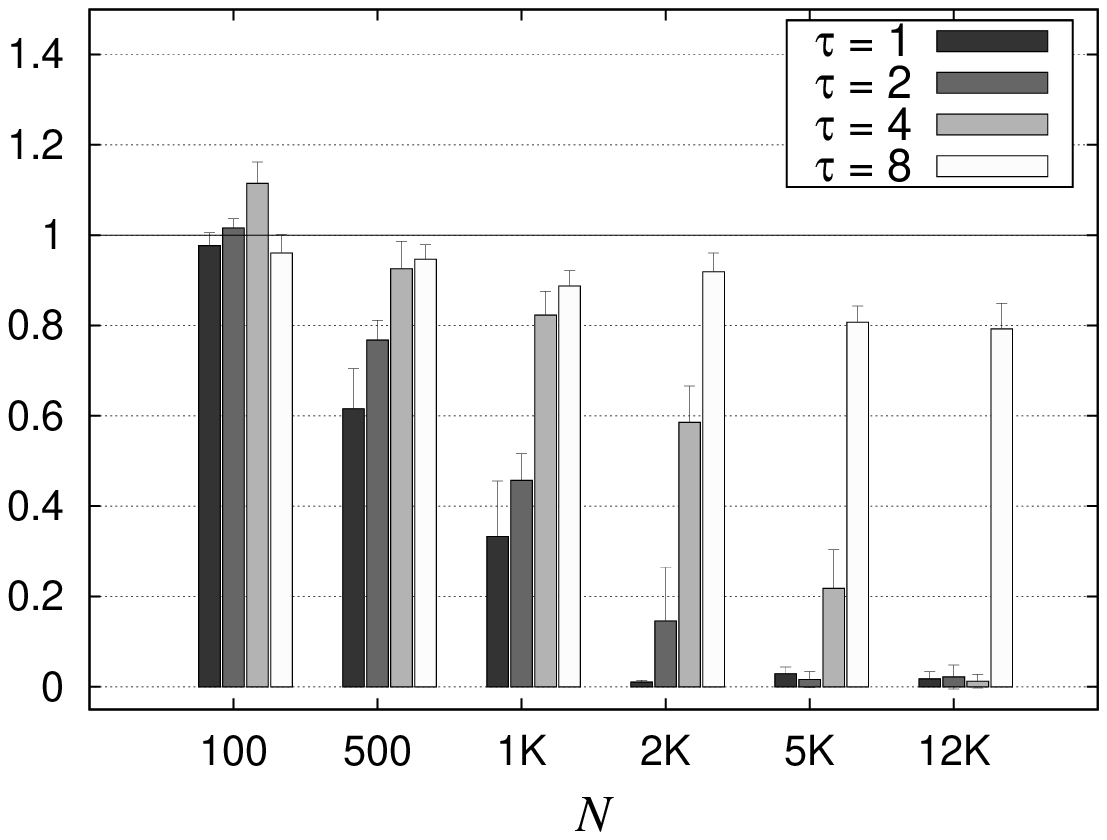,width=6cm} \\ \hline
    \end{tabular}
   \caption{\label{fig:c_h} $\rHe{HC}$ (left) and $\rHitrue{HC}$ (right) for 
   datasets with $n=50$ and $\tau=1,2,4,8$ (color-coded) ran on datasets with increasing
   number of datapoints $N$ . Bars represents average over
   ten different datasets. Error bars represents standard deviation.}  
\end{figure}

This tendency is clearer in the results for $n=50$, shown in Table \ref{table:c},
plotted in Figure~\ref{fig:c_h}. The
table shows the mean value and standard deviation (in parenthesis) of edge Hamming distance
$\He{GSMN}$, $\He{HC}$ and their ratio $\rHe{HC}$ on the left group of columns, and $\Hitrue{GSMN}$,
$\Hitrue{HC}$ and their ratio  $\rHitrue{HC}$ on the right group of columns. Again, bold
(underline) signifies an increase (decrease) in the quality IBMAP-HC over the quality of
GSMN. We can observe that for the case of edge Hamming distance,  as $\tau$ increases,
the number of bold ratios decreases ($5$,$4$,$2$, and $2$ for $\tau=1,2,4,8$, respectively), and
the number of underlined ratios increases ($1$,$2$,$3$, and $2$ for $\tau=1,2,4,8$,
respectively). 

The plots in Figure~\ref{fig:c_h} show clearly how both ratios $\rHe{HC}$ (left) and
$\rHitrue{HC}$ (right) decrease with $N$, with the decrease being slower as $\tau$
increases. Although for small $N$s there are some pour results for edge errors, these
ratios are never greater than $1.231$ (for $\tau=4$, $N=500$), and in most
cases has a corresponding improvement in the independence error. For instance, for the
same case of $\tau=4$, $N=500$, the independence ratio is $0.926 (0.061)$, a value
smaller than $1$ with statistical significance.  In all other cases, only the four cases
of $N=100$ show no improvement, with only the case of $\tau=4$ showing an increase in
error. The remaining cases of edge errors show significant improvements reaching in many
cases of large $N$s ratios smaller than $0.1$, with proportions of $50$ to $1$ wrong
edges in GSMN and IBMAP-HC, respectively, for $\tau=4$, $N=12000$.

\begin{table}[h]
\renewcommand{\tabcolsep}{5pt}
\centering
\begin{scriptsize}
\begin{tabular}{|c|c||c|c|c||c|c|c|}
\hline
\multicolumn{8}{|c|}{$n=50$} \\ \hline
 &  & \multicolumn{3}{c||}{ } & \multicolumn{3}{|c|}{ } \\
 &  & \multicolumn{3}{c||}{Edges Hamming distances} & \multicolumn{3}{|c|}{Independences Hamming distances} \\
 &  & \multicolumn{3}{c||}{ } & \multicolumn{3}{|c|}{ } \\ \cline{3-8}
& & & & & & & \\
$\tau$ & $N$ & $\He{GSMN}$ & $\He{HC}$ & $\rHe{HC}$ & $\Hitrue{GSMN}$ & $\Hitrue{HC}$ & $\rHitrue{HC}$ \\
& & & & & & & \\
\hline

& & & & & & & \\
\multirow{6}{*}{1} & 100 & 193.200(4.255) & 205.800(3.133) & \underline{1.066(0.026)} & 0.789(0.010) & 0.770(0.022) & 0.977(0.029)  \\
 & 500 & 156.500(9.919) & 145.200(16.865) & \textbf{ 0.923(0.063)} & 0.720(0.023) & 0.447(0.075) & \textbf{0.616(0.089)}  \\
 & 1000 & 91.200(10.771) & 37.300(18.338) & \textbf{0.389(0.160)} & 0.556(0.034) & 0.189(0.072) & \textbf{0.333(0.122)}  \\
 & 2000 & 55.000(12.296) & 1.556(0.422) & \textbf{0.030(0.009)} & 0.389(0.075) & 0.004(0.002) & \textbf{0.010(0.004)}  \\
 & 5000 & 22.200(3.758) & 1.700(0.943) & \textbf{0.071(0.040)} & 0.163(0.034) & 0.005(0.003) & \textbf{0.029(0.015)}  \\
 & 12000 & 13.400(2.514) & 0.800(0.728) & \textbf{0.055(0.049)} & 0.098(0.020) & 0.002(0.002) & \textbf{0.018(0.016)}  \\
& & & & & & & \\
\hline

& & & & & & & \\
\multirow{6}{*}{2} & 100 & 186.200(4.370) & 206.200(3.959) & \underline{1.108(0.025)} & 0.605(0.014) & 0.615(0.020) & 1.016(0.020)  \\
 & 500 & 153.500(7.465) & 163.900(12.135) & \underline{1.067(0.055)} & 0.566(0.022) & 0.435(0.036) &\textbf{ 0.767(0.044)}  \\
 & 1000 & 117.100(11.527) & 93.200(21.019) & \textbf{0.778(0.108)} & 0.502(0.027) & 0.230(0.035) & \textbf{0.457(0.059)}  \\
 & 2000 & 81.400(10.540) & 14.500(13.967) & \textbf{0.150(0.137)} & 0.433(0.030) & 0.067(0.055) & \textbf{0.146(0.119)}  \\
 & 5000 & 35.700(4.595) & 0.900(0.966) & \textbf{0.024(0.025)} &  0.273(0.028) & 0.005(0.005) & \textbf{0.016(0.017)}  \\
 & 12000 & 20.600(3.328) & 0.600(0.595) & \textbf{0.034(0.037)} & 0.163(0.019) & 0.003(0.004) & \textbf{0.022(0.026)}  \\
& & & & & & & \\
\hline
& & & & & & & \\
\multirow{6}{*}{4}  & 100 & 167.727(9.113) & 199.818(6.557) & \underline{ 1.195(0.033)} & 0.257(0.012) & 0.286(0.012) & \underline{1.115(0.047)}  \\
 & 500 & 128.700(5.097) & 158.200(7.175) & \underline{1.231(0.056)} & 0.258(0.012) & 0.238(0.016)  & \textbf{0.926(0.061)}  \\
 & 1000 & 118.900(4.315) & 140.900(5.791) & \underline{1.186(0.034)} & 0.243(0.009) & 0.200(0.014) & \textbf{0.823(0.053)}  \\
 & 2000 & 106.200(6.297) & 98.900(12.825) & 0.928(0.089) & 0.219(0.008) & 0.129(0.021) & \textbf{0.585(0.081)}  \\
 & 5000 & 85.800(7.594) & 35.100(14.526) & \textbf{0.395(0.142)} & 0.205(0.012) & 0.045(0.018)  & \textbf{0.218(0.085)}  \\
 & 12000 & 51.800(6.200) & 0.900(1.348) & \textbf{0.015(0.022)} & 0.145(0.011) & 0.002(0.002)  & \textbf{0.012(0.015)}  \\
& & & & & & & \\
\hline
& & & & & & & \\
\multirow{6}{*}{8}  & 100 & 296.000(7.433) & 298.600(9.795) & 1.009(0.024) & 
0.157(0.004) & 0.150(0.006)  & 0.961(0.040)  \\
 & 500 & 261.900(4.707) & 276.700(4.760) & \underline{1.057(0.024)} & 0.137(0.003) & 0.130(0.004) & \textbf{0.947(0.032)}  \\
 & 1000 & 237.800(9.145) & 255.700(8.231) & \underline{1.077(0.042)} & 0.134(0.004) & 0.119(0.002) & \textbf{0.888(0.033)}  \\
 & 2000 & 244.222(13.415) & 253.889(10.707) & 1.042(0.045) & 0.125(0.005) & 0.115(0.005) & \textbf{0.919(0.041)}  \\
 & 5000 & 263.000(19.221) & 225.000(9.314) & \textbf{0.860(0.047)} & 0.127(0.006) & 0.102(0.003) & \textbf{0.807(0.036)}  \\
 & 12000 & 232.100(23.718) & 188.400(12.461) & \textbf{0.822(0.071}) & 0.112(0.007) & 0.089(0.005) & \textbf{0.793(0.056)}  \\
& & & & & & & \\
\hline
\end{tabular}
\caption{\label{table:c} Experiments results of edge (left) and independence (right) Hamming distances
for several datasets with $ n = 50 $, 
$\tau=1,2,4,8$ and increasing number of data points $N$ (rows). Standard deviations are
shown in parenthesis. Again, quality improvements are in bold, and quality reductions are
underlined. }   
\end{scriptsize}
\end{table}


\begin{figure}[h]
   \centering
   \begin{tabular}{c|c} 
	\multicolumn{2}{c}{Growth in the number of ascents of IBMAP-HC} \\ \hline
 	\epsfig{figure=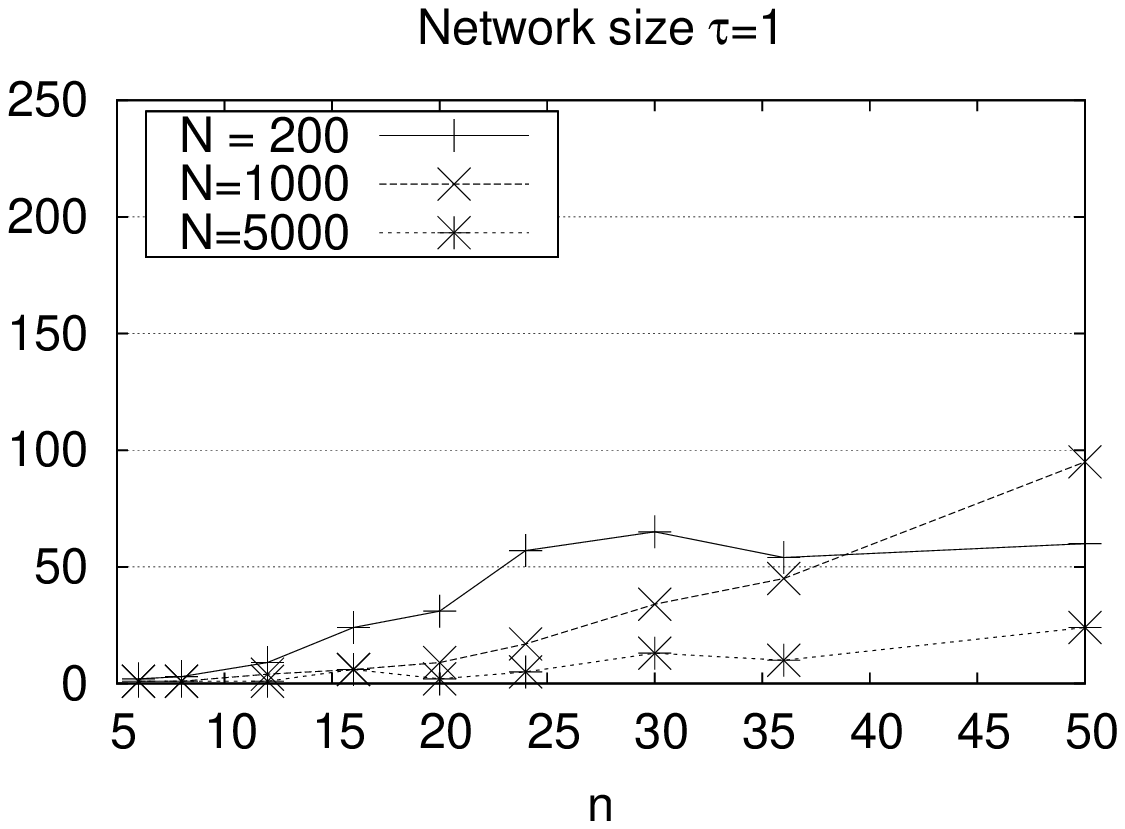,width=7cm} & 
	\epsfig{figure=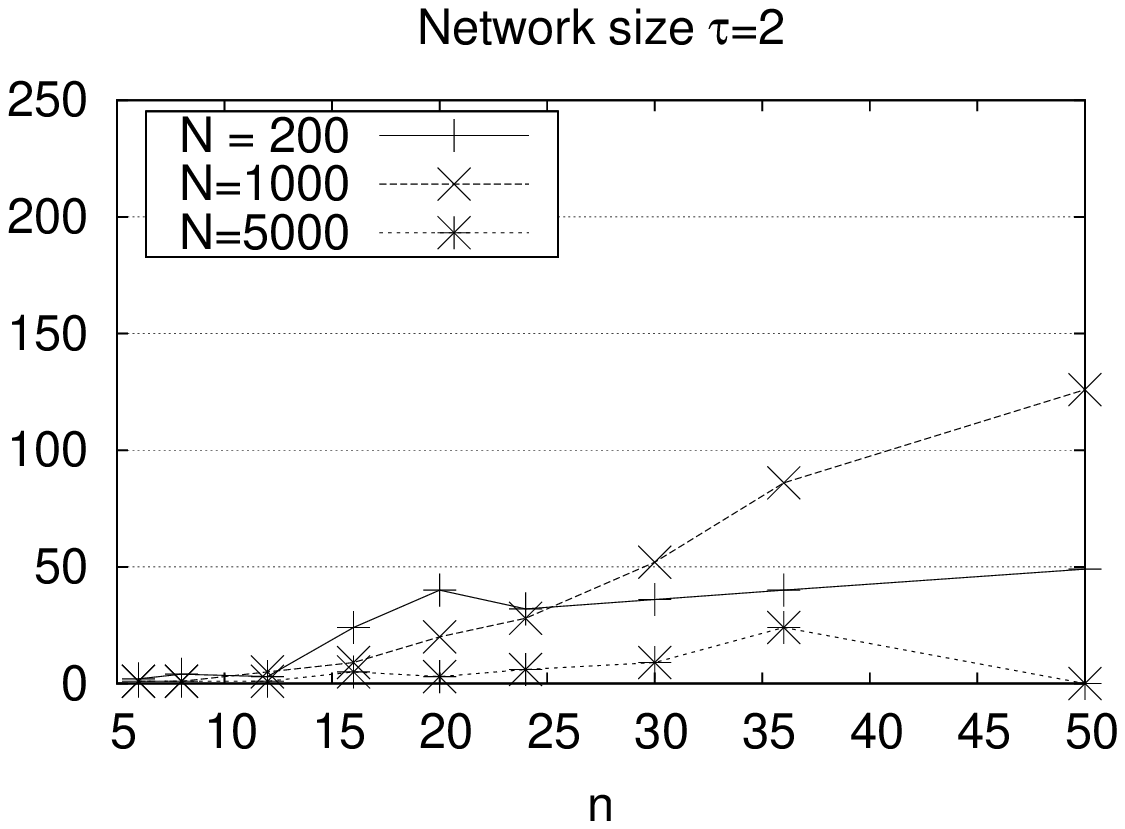,width=7cm} \\ 
 	\epsfig{figure=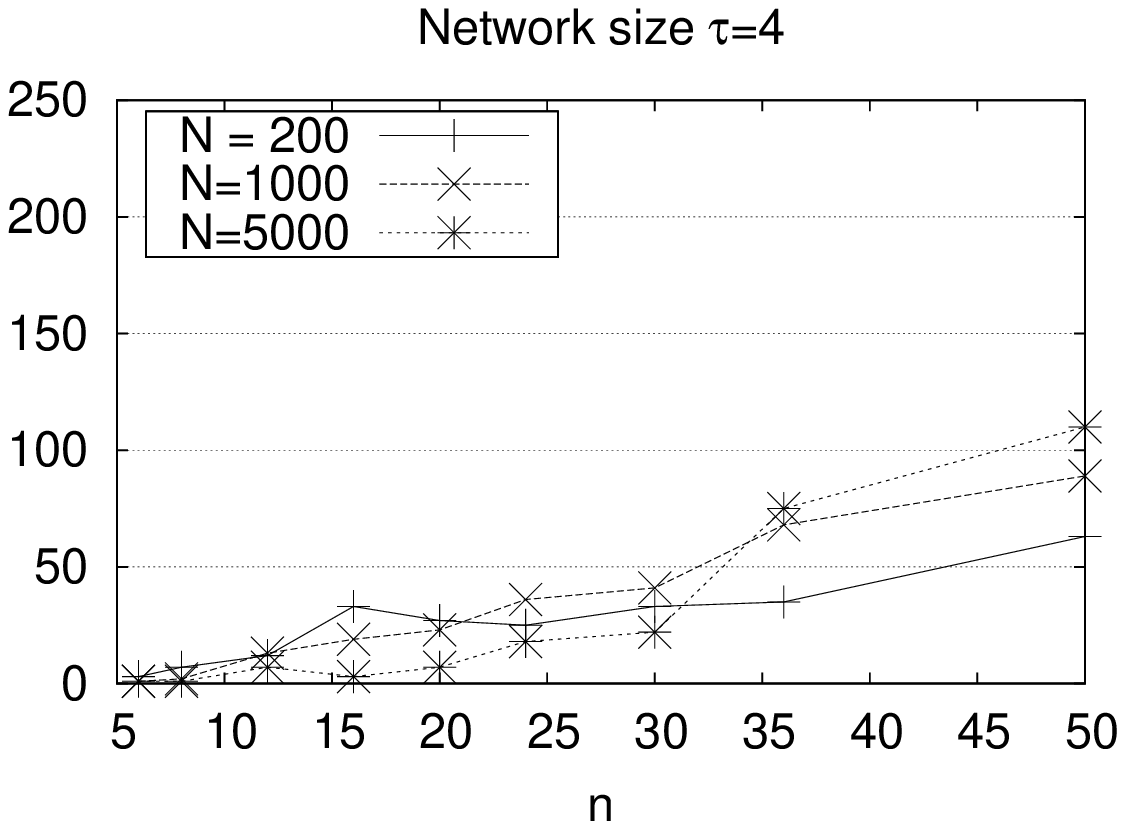,width=7cm} & 
	\epsfig{figure=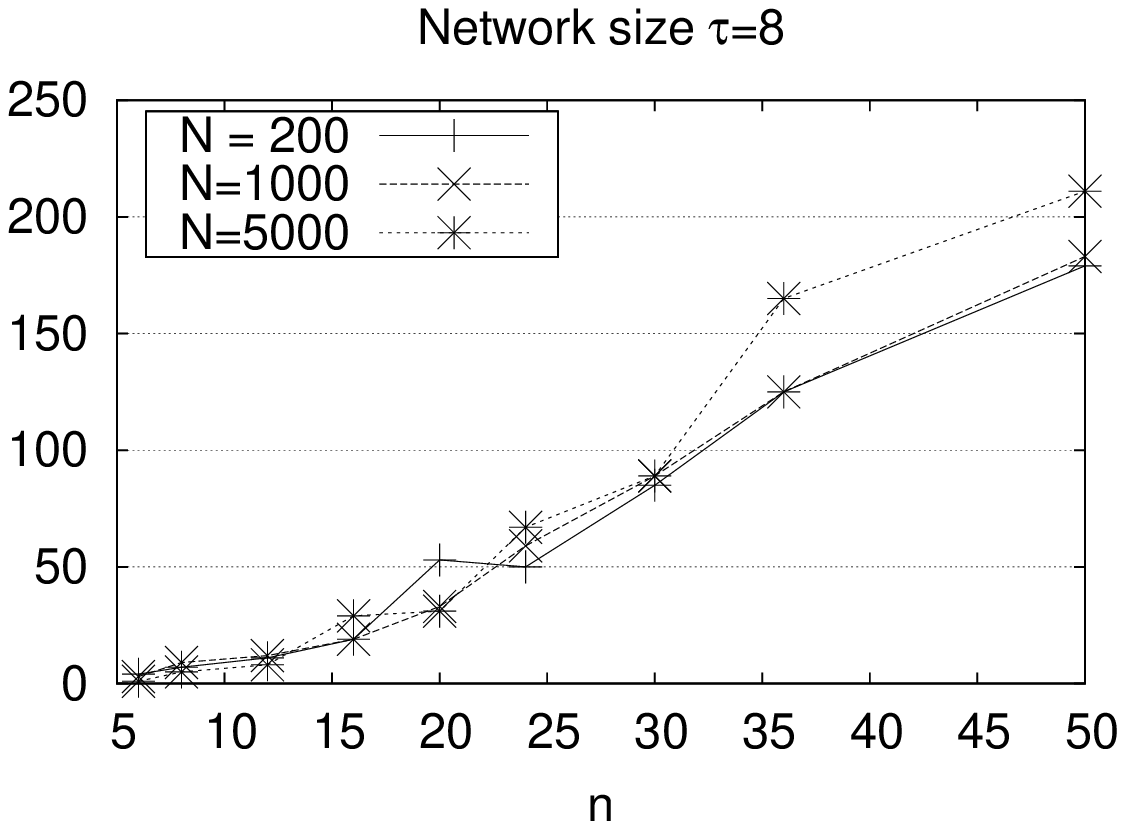,width=7cm} \\ \hline
    \end{tabular}
   \caption{ \label{fig:wnt} Number of IBMAP-HC ascents $M$ on datasets with increasing
   number of variables $n$, connectivities $\tau=1,2,4,8$ (four plots), $N=200, 1000,
   5000$ (three curves).}
 \end{figure}

To conclude this section we present results for $M$, the number of hill climbs conducted
by IBMAP-HC. We measured this quantity for
runs of the algorithms on datasets sampled from graphs of increasing size $n$,  
connectivities $\tau=1,2,4,8$, and three different reliability conditions
$N=200,1000,5000$. Figure \ref{fig:wnt} shows four plots, one per $\tau$, with $3$ curves each,
one per dataset size $N$. 

In all cases, $M$ presents a quasi-linear grow on $n$. After incorporating $M$ as a
a linear function on $n$ into the expression $O(Mn(n-1)^2 N \tau^*) $ for the complexity
of the IBMAP-HC algorithm (c.f. \S \ref{sec:complexity}), we obtain empirically the
runtime grows as $O(n^4)$. 

To give the reader a sense of the runtime, we report the runtime for the hardest
scenario. In a Java virtual machine running on an AMD
Athlon processor of 64 bits and 4800 Mhz and 1 Gb of RAM, the case for $n=50$, $N=12000$,
and $\tau=8$ took 5 days, with a reduction to 5 hours after implementing a simple cache
for avoiding the computation of repeated tests.

\subsubsection{Experiments on benchmark datasets}

To conclude the discussion of our experiments we compare quality performances of IBMAP-HC against
GSMN on benchmark datasets.  For this datasets the underlying true network is unknown,
and we are thus restricted to independence Hamming distances measured on the dataset. 

We ran the IBMAP-HC and GSMN algorithms ten times on each of a set of datasets obtained from the UCI
Repositories of machine learning \citep{Asuncion+Newman:2007} and KDD datasets
\citep{Hettich+Bay:1999}. Each ran on a different randomly sampled subset of the whole
dataset of $1/3$ its size. The same was repeated for subsets of $1/5$ the total size. The
independence Hamming distances $\Hidata{HC}$, and $\Hidata{GSMN}$ where 
then computed comparing the independencies of the output networks with the independencies
of the complete dataset. 

\begin{figure}
   \centering
   \begin{tabular}{c|c} 
	\multicolumn{2}{c}{Comparison of independence Hamming distances }\\
	\multicolumn{2}{c}{for benchmark data sets}\\ \hline
	$N/3$ & $N/5$ \\
 	\epsfig{figure=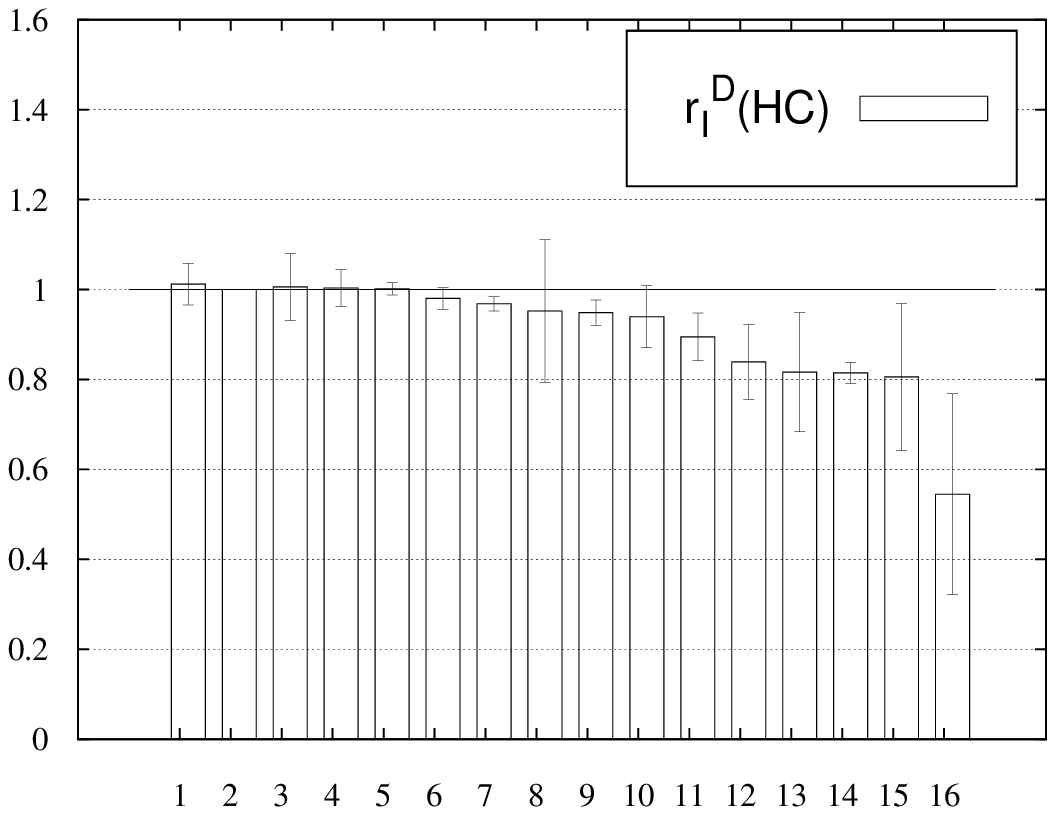,width=7cm} & \epsfig{figure=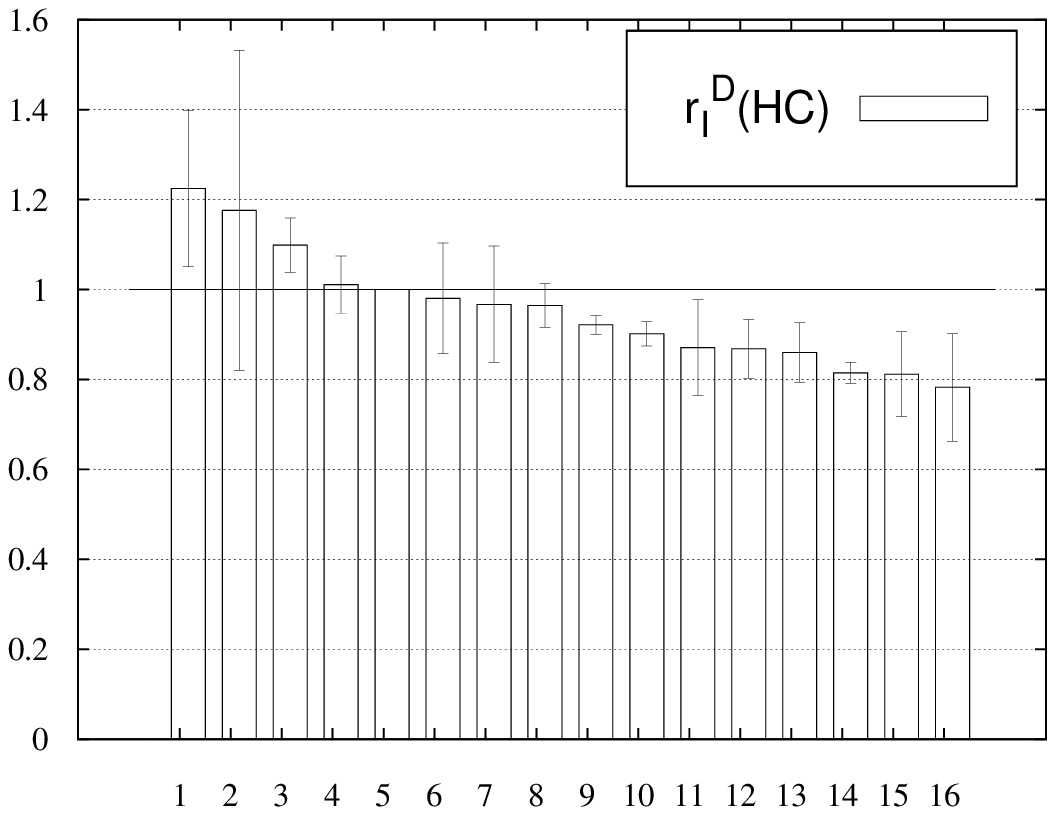,width=7cm} \\ 
	& \\ \hline
    \end{tabular}
   \caption{\label{fig:de} Experimental results for real datasets. The table shows
   average and standard  deviation of    $\rHidata{HC}$ over $10$, 
   $1/3$ subsets of $\data$ (left) and over $10$, $1/5$ subsets (right).
   Again, smaller values mean greater improvements. }   
\end{figure}

Results of these experiments are shown in Table \ref{table:d} and Figure
\ref{fig:de} (left) for the $1/3$ case, and Table \ref{table:e} and
Figure \ref{fig:de} (right) for the $1/5$ case. On both figures, the numbers in the
$x$-axis represent the indexes of each dataset in the corresponding table. The tables also show on
their last two columns values for $\rHidata{HC}$ and $M$. These results show improvements in
most cases, again, shown in bold in the tables and bars lower than one in the figures.
Quality reductions are underlined in both tables.

These results show again quality improvements of IBMAP-HC over GSMN, with only two cases,
lenses and flare2 in the $1/5$ case showing reductions. Improvements occurred in the $8$
out of $16$ in both cases.

\begin{table}
\begin{scriptsize}
\centering
\begin{tabular}{|c|ccc|c c|c|c|}
\hline
& & & & & & & \\
\# & name & n & N & $\Hidata{GSMN}$ & $\Hidata{HC}$ & $\rHidata{HC}$ & $M$\\
& & & & & & & \\
\hline
& & & & & & & \\
1 & bridges.csv & 12 & 140 & 0.445(0.018) & 0.450(0.020) & 1.012(0.046) & 7.700(1.105)\\ 
2 & dependent.csv & 3 & 1000 & 0.000(0.000) & 0.000(0.000) & 1.000(0.000) & 1.000(0.000)\\ 
3 & flare2.csv & 13 & 1065 & 0.203(0.009) & 0.204(0.016) & 1.006(0.074) & 10.200(2.298)\\ 
4 & haberman.csv & 5 & 612 & 0.309(0.029) & 0.309(0.024) & 1.003(0.040) & 1.500(0.372)\\ 
5 & imports-85.csv & 25 & 386 & 0.601(0.006) & 0.602(0.007) & 1.001(0.014) & 20.667(1.942)\\ 
6 & balance-scale.csv & 5 & 1250 & 0.421(0.015) & 0.412(0.013) & 0.980(0.024) & 1.100(0.223)\\ 
7 & car.csv & 7 & 3456 & 0.425(0.006) & 0.412(0.006) & \textbf{0.968(0.016)} & 2.000(0.412)\\ 
8 & monks-1.csv & 7 & 1112 & 0.091(0.011) & 0.091(0.031) & 0.952(0.158) & 1.182(0.393)\\ 
9 & dermatology.csv & 35 & 357 & 0.724(0.008) & 0.687(0.025) & \textbf{0.949(0.028)} & 69.556(6.525)\\ 
10 & glass.csv & 10 & 210 & 0.449(0.029) & 0.418(0.014) & 0.939(0.069) & 5.700(1.054)\\ 
11 & nursery.csv & 9 & 25920 & 0.551(0.035) & 0.491(0.024) & \textbf{0.895(0.053)} & 4.000(0.940)\\ 
12 & machine.csv & 10 & 201 & 0.411(0.044) & 0.341(0.015) & \textbf{0.839(0.083)} & 6.750(1.088)\\ 
13 & hayes-roth.csv & 6 & 264 & 0.358(0.056) & 0.291(0.061) & \textbf{0.816(0.133)} & 3.600(1.531)\\ 
14 & tictactoe.csv & 10 & 1916 & 0.584(0.021) & 0.476(0.021) & \textbf{0.815(0.024)} & 9.500(1.459)\\ 
15 & baloons.csv & 5 & 20 & 0.324(0.113) & 0.244(0.087) & \textbf{0.806(0.164)} & 1.900(0.617)\\ 
16 & lenses.csv & 5 & 48 & 0.306(0.080) & 0.145(0.046) & \textbf{0.545(0.223)} & 2.200(0.446)\\ 
& & & & & & & \\
\hline
\end{tabular}
\caption{\label{table:d}Real data independence Hamming errors experiments for a $1/3$ of the complete
dataset. Quality improvements when $r<1$ (in bold), and quality reductions when $r>1$ (underlined). } 
\end{scriptsize}
\end{table}

\begin{table}
\begin{scriptsize}
\centering
\begin{tabular}{|c|ccc|c c|c|c|}
\hline
& & & & & & & \\
\# & name & n & N & $\Hidata{GSMN}$ & $\Hidata{HC}$ & $\rHidata{HC}$ & $M$\\
& & & & & & & \\
\hline
& & & & & & & \\
1 & lenses.csv & 5 & 48 & 0.241(0.076) & 0.294(0.093) & \underline{1.225(0.174)} & 1.700(0.341)\\ 
2 & baloons.csv & 5 & 20 & 0.378(0.407) & 0.417(0.406) & 1.176(0.356) & 1.100(0.844)\\ 
3 & flare2.csv & 13 & 1065 & 0.190(0.004) & 0.208(0.011) & \underline{1.099(0.060)} & 9.300(1.792)\\ 
4 & bridges.csv & 12 & 140 & 0.404(0.022) & 0.407(0.027) & 1.011(0.063) & 9.000(1.287)\\ 
5 & dependent.csv & 3 & 1000 & 0.000(0.000) & 0.000(0.000) & 1.000(0.000) & 1.000(0.000)\\ 
6 & glass.csv & 10 & 210 & 0.358(0.041) & 0.342(0.015) & 0.980(0.123) & 7.300(0.943)\\ 
7 & machine.csv & 10 & 201 & 0.357(0.047) & 0.336(0.024) & 0.967(0.130) & 5.400(0.595)\\ 
8 & haberman.csv & 5 & 612 & 0.315(0.023) & 0.302(0.019) & 0.965(0.049) & 1.900(0.617)\\ 
9 & dermatology.csv & 35 & 357 & 0.709(0.006) & 0.653(0.012) & \textbf{0.922(0.021)} & 58.100(4.490)\\ 
10 & imports-85.csv & 25 & 386 & 0.550(0.024) & 0.495(0.016) & \textbf{0.902(0.027)} & 19.900(2.290)\\ 
11 & balance-scale.csv & 5 & 1250 & 0.422(0.033) & 0.366(0.048) & \textbf{0.871(0.107)} & 1.700(0.581)\\ 
12 & nursery.csv & 9 & 25920 & 0.469(0.029) & 0.405(0.028) & \textbf{0.868(0.066)} & 4.100(0.966)\\ 
13 & car.csv & 7 & 3456 & 0.386(0.034) & 0.331(0.040) & \textbf{0.860(0.066)} & 3.000(0.940)\\ 
14 & tictactoe.csv & 10 & 1916 & 0.584(0.021) & 0.476(0.021) & \textbf{0.815(0.024)} & 9.500(1.459)\\ 
15 & monks-1.csv & 7 & 1112 & 0.126(0.041) & 0.096(0.018) & \textbf{0.812(0.094)} & 1.300(0.341)\\ 
16 & hayes-roth.csv & 6 & 264 & 0.302(0.062) & 0.227(0.040) & \textbf{0.782(0.120)} & 1.900(0.844)\\ 
& & & & & & & \\
\hline
\end{tabular}
\caption{\label{table:e}Real data independence Hamming errors experiments for a $1/5$ of the complete
dataset. Quality improvements when $r<1$ (in bold), and quality reductions when $r>1$ (underlined).}  
\end{scriptsize}
\end{table}

\section{Conclusions and future work\label{sec:conclusions}} 
In conclusion, the IBMAP-HC algorithm provides the possibility of solving the problem for
complex systems getting significant quality improvements over GSMN. IB-score seems to be
a good and efficient likelihood function.  

From this work, several future research possibilities arise that we are motivated to
pursue, including 
continuing to look into a practical algorithm with lower computational cost (to be used
with even more complex systems), experience with other approximate optimization
algorithms such as  local beam Metropolis Hastings, continue with the analysis of quality
measures and propose new, more efficient and reliable scoring 
functions, and perform a thorough comparison of these novel scoring functions and
existing scoring functions such as the likelihood of the data given the complete model
(structure plus parameters).      

\section{Acknowledgements}	
This work was funded by a postdoctoral fellowship from CONICET, Argentina; and the
scholarship program for teachers of the National Technological University,
Ministry of Science, Technology and
Productive Innovation and the National Agency of Scientific and Technological Promotion,
FONCyT; Argentina.    

\appendix 
\section{Proof of Lemma \ref{thm:blanketDefOpp} \label{app:app1}}

All along this work we made the running assumption of graph-isomorphism, that is, that
the underlying distribution we are trying to learn has a graph that encodes all and only the
independencies that holds in the distribution. According to Theorem 2 of
[\citet{pearl88}, p.97], the following properties among independence  
hold in any graph-isomorph distribution:

\begin{itemize}
 \item Intersection: $\ci{X}{Y}{\set{Z}, W} \wedge \ci{X}{W}{\set{Z}, Y} \Rightarrow \ensuremath{I({X};{Y,W} \mid \set{Z})}$
 \item Decomposition: $\ensuremath{I({X};{Y,W} \mid \set{Z})} \Rightarrow \ci{X}{Y}{\set{Z}} \wedge \ci{X}{W}{\set{Z}}$
 \item Strong Union: $\ci{X}{Y}{\set{Z}} \Rightarrow \ci{X}{Y}{\set{Z},W}$
\end{itemize}


Before proving the main lemma of the appendix we present an auxiliaty lemma.

\begin{auxiliarylemma}\label{lemma:AL1}
	For all $X \neq Y \neq W \in \set{V}$ and $\set{Z} \subseteq \set{V}-\{X,Y,W\}$, 
 	\begin{equation*} 
		\cd{X}{Y}{\set{Z}} \wedge \ci{X}{Y}{{\set{Z}, W}} \Rightarrow \cd{X}{W}{\set{Z}} 
	\end{equation*}
\end{auxiliarylemma}

\begin{proofJMLR}
		By Intersection and decomposition, 
		 	\begin{eqnarray*} 
				\ci{X}{Y}{\set{Z},W} \wedge \ci{X}{W}{\set{Z},Y} &\Rightarrow& I(X;Y,W \mid
				\set{Z}) \\
										&\Rightarrow& \ci{X}{Y}{\set{Z}}. 
			\end{eqnarray*}
		Then, by the counter-positive of this expression and the counter-positive of
		Strong Union, 
		 	\begin{eqnarray*} 
		\cd{X}{Y}{\set{Z}} \wedge \ci{X}{Y}{{\set{Z}, W}} &\Rightarrow& \cd{X}{W}{\set{Z},
		Y} \\
														&\Rightarrow& \cd{X}{W}{\set{Z}}. 
			\end{eqnarray*}
\end{proofJMLR}


We can now prove Lemma \ref{thm:blanketDefOpp}, reproduced here for convenience:
\newtheorem*{LemmaBlanketDefOpp}{Lemma \ref{thm:blanketDefOpp}}
\begin{LemmaBlanketDefOpp}
For every $W\neq X \in \set{V}$, and $\blanket{X}$ the boundary of $X$, 
 	\begin{equation*} 
		W \notin \blanket{X} \Leftarrow \ci{X}{W}{{\blanket{X}-\{X,W\}}} 
	\end{equation*}
\end{LemmaBlanketDefOpp}

\begin{proofJMLR}

	    We prove its counter-positive $W \in \blanket{X} \Rightarrow \cd{X}{W}{{\blanket{X}-\{X,W\}}} $.
		By minimality of $\blanket{X}$, there must be a variable $Y \notin \blanket{X}$
		such that removing $W$ from the blanket, $Y$ becomes dependent of $X$, \ie, $~\cd{X}{Y}{{\blanket{X}-\{W\}-\{X,Y\}}} $. Also, since $Y \notin \blanket{X}$, by the
		definition of boundary of Eq.~(\ref{eq:blanketDef}) it holds that
		$\ci{X}{Y}{{\blanket{X}-\{X,Y\}}}$. The lemma follows then from Auxiliary Lemma
		\ref{lemma:AL1} by letting $\set{Z} = \blanket{X}-\{X,Y\} - \{W\}$.
\end{proofJMLR}
	
\clearpage
\bibliography{JMLR2010-BROMBERG-SCHLUTER-IB-SCORE}

\begin{thebibliography}{26}
\providecommand{\natexlab}[1]{#1}
\providecommand{\url}[1]{\texttt{#1}}
\expandafter\ifx\csname urlstyle\endcsname\relax
  \providecommand{\doi}[1]{doi: #1}\else
  \providecommand{\doi}{doi: \begingroup \urlstyle{rm}\Url}\fi

\bibitem[A.~Asuncion(2007)]{Asuncion+Newman:2007}
D.J.~Newman A.~Asuncion.
\newblock {UCI} machine learning repository, 2007.

\bibitem[Acid and {de Campos}(2003)]{ACID03}
S.~Acid and Luis~M. {de Campos}.
\newblock Searching for {Bayesian} network structures in the space of
  restricted acyclic partially directed graphs.
\newblock \emph{Journal of Artificial Intelligence Research}, 18:\penalty0
  445--490, 2003.

\bibitem[Agresti(2002)]{AGRESTI02}
Alan Agresti.
\newblock \emph{Categorical Data Analysis}.
\newblock Wiley, 2nd edition, 2002.

\bibitem[Aliferis et~al.(2003)Aliferis, Tsamardinos, and Statnikov]{HITON03}
C.~F. Aliferis, I.~Tsamardinos, and A.~Statnikov.
\newblock {HITON}, a novel {Markov} blanket algorithm for optimal variable
  selection.
\newblock In \emph{Proceedings of the American Medical Informatics Association
  (AMIA) Fall Symposium}, 2003.

\bibitem[Anguelov et~al.(2005)Anguelov, Taskar, Chatalbashev, Koller, Gupta,
  Heitz, and Ng]{ANGUELOV&TASKAR05}
D.~Anguelov, B.~Taskar, V.~Chatalbashev, D.~Koller, D.~Gupta, G.~Heitz, and
  A.~Ng.
\newblock Discriminative learning of {Markov} random fields for segmentation of
  {3D} range data.
\newblock \emph{Proceedings of the Conference on Computer Vision and Pattern
  Recognition}, 2005.

\bibitem[Besag(1974)]{BESAG74}
J.~Besag.
\newblock Spacial interaction and the statistical analysis of lattice systems.
\newblock \emph{Journal of the Royal Statistical Society, Series B}, 1974.

\bibitem[Besag(1975)]{BESAG75}
J.~Besag.
\newblock Statistical analysis of non-lattice data.
\newblock \emph{The Statistician}, 24\penalty0 (3):\penalty0 179--195, 1975.

\bibitem[Besag et~al.(1991)Besag, York, and Mollie]{BESAG91}
J.~Besag, J.~York, and A.~Mollie.
\newblock {Bayesian} image restoration with two applications in spatial
  statistics.
\newblock \emph{Annals of the Inst. of Stat. Math.}, 43:\penalty0 1--59, 1991.

\bibitem[Bromberg and Margaritis(2009)]{BrombMarg09}
F.~Bromberg and D.~Margaritis.
\newblock Improving the reliability of causal discovery from small data sets
  using argumentation.
\newblock \emph{Journal of Machine Learning Research}, 10:\penalty0 301--340,
  Feb 2009.

\bibitem[Bromberg et~al.(2009)Bromberg, Margaritis, and
  V.]{bromberg&margaritis09b}
F.~Bromberg, D.~Margaritis, and Honavar V.
\newblock Efficient markov network structure discovery using independence
  tests.
\newblock \emph{Journal of Artificial Intelligence Research}, 35:\penalty0
  449--485, July 2009.

\bibitem[Cochran(1954)]{COCHRAN54}
W.~G. Cochran.
\newblock Some methods of strengthening the common ${\chi}$ tests.
\newblock \emph{Biometrics.}, page 10:417–451, 1954.

\bibitem[Friedman et~al.(2000)Friedman, Linial, Nachman, and Pe'er]{friedman00}
N.~Friedman, M.~Linial, I.~Nachman, and D.~Pe'er.
\newblock Using {Bayesian} networks to analyze expression data.
\newblock \emph{Computational Biology}, 7:\penalty0 601--620, 2000.

\bibitem[Heckerman et~al.(1995)Heckerman, Geiger, and Chickering]{Heckerman95}
D.~Heckerman, D.~Geiger, and D.~M. Chickering.
\newblock Learning {Bayesian} networks: The combination of knowledge and
  statistical data.
\newblock \emph{Machine Learning}, 1995.

\bibitem[Hettich and Bay(1999)]{Hettich+Bay:1999}
S.~Hettich and S.~D. Bay.
\newblock The {UCI KDD} archive, 1999.
\newblock URL \url{http://kdd.ics.uci.edu}.

\bibitem[Koller and Friedman(2009)]{KOLLER&FRIEDMAN09}
D.~Koller and N.~Friedman.
\newblock \emph{Probabilistic Graphical Models: Principles and Techniques}.
\newblock MIT Press, Cambridge, MA, 2009.

\bibitem[Lam and Bacchus(1994)]{LamBacchus94}
W.~Lam and F.~Bacchus.
\newblock Learning {Bayesian} belief networks: an approach based on the {MDL}
  principle.
\newblock \emph{Computational Intelligence}, 10:\penalty0 269--293, 1994.

\bibitem[Lauritzen(1996)]{LAURITZEN96}
S.~L. Lauritzen.
\newblock \emph{Graphical Models}.
\newblock Oxford University Press, 1996.

\bibitem[Margaritis(2005)]{MARGARITIS05}
D.~Margaritis.
\newblock Distribution-free learning of {Bayesian} network structure in
  continuous domains.
\newblock In \emph{Proceedings of AAAI}, 2005.

\bibitem[Margaritis and Bromberg(2009)]{margaritisBromberg09}
D.~Margaritis and F.~Bromberg.
\newblock Efficient markov network discovery using particle filter.
\newblock \emph{Computational Intelligence}, 25\penalty0 (4):\penalty0
  367--394, September 2009.

\bibitem[Margaritis and Thrun(2000)]{MARGARITIS00}
D.~Margaritis and S.~Thrun.
\newblock {Bayesian} network induction via local neighborhoods.
\newblock In S.A. Solla, T.K. Leen, and K.R. M{\"u}ller, editors,
  \emph{Advances in Neural Information Processing Systems 12}, pages 505--511.
  MIT Press, 2000.

\bibitem[McCallum(2003)]{MCCALLUM03}
A.~McCallum.
\newblock Efficiently inducing features of conditional random fields.
\newblock In \emph{Proceedings of Uncertainty in Artificial Intelligence
  (UAI)}, 2003.

\bibitem[Pearl(1988)]{pearl88}
J.~Pearl.
\newblock \emph{Probabilistic Reasoning in Intelligent Systems: Networks of
  Plausible Inference}.
\newblock Morgan Kaufmann Publishers, Inc., 1988.

\bibitem[Shekhar et~al.(2004)Shekhar, Zhang, Huang, and Vatsavai]{shekhar04}
S.~Shekhar, P.~Zhang, Y.~Huang, and R.~R. Vatsavai.
\newblock In H.~Kargupta, A.~Joshi, K.~Sivakumar, and Y.~Yesha, editors,
  \emph{Trends in Spatial Data Mining}, chapter~19, pages 357--379. AAAI Press
  / The MIT Press, 2004.

\bibitem[Spirtes et~al.(2000)Spirtes, Glymour, and Scheines]{Spirtes00}
P.~Spirtes, C.~Glymour, and R.~Scheines.
\newblock \emph{Causation, Prediction, and Search}.
\newblock Adaptive Computation and Machine Learning Series. MIT Press, 2000.

\bibitem[Tsamardinos et~al.(2003)Tsamardinos, Aliferis, and Statnikov]{IAMB03}
I.~Tsamardinos, C.~F. Aliferis, and A.~Statnikov.
\newblock Algorithms for large scale {Markov} blanket discovery.
\newblock In \emph{Proceedings of the 16th International FLAIRS Conference},
  pages 376--381, 2003.

\bibitem[Yu and Cheng(2003)]{YU-CHENG2003}
Y.~Yu and Q.~Cheng.
\newblock {MRF} parameter estimation by an accelerated method.
\newblock \emph{Pattern Recognition Letters}, 24\penalty0 (9-10):\penalty0
  1251--1259, 2003.

\end{thebibliography}

\end{document}